\documentclass[sigconf]{acmart}

\AtBeginDocument{%
  \providecommand\BibTeX{{%
    \normalfont B\kern-0.5em{\scshape i\kern-0.25em b}\kern-0.8em\TeX}}}

\copyrightyear{2021}
\acmYear{2021}
\setcopyright{acmcopyright}
\acmConference[RecSys '21]{Fifteenth ACM Conference on Recommender Systems}{September 27-October 1, 2021}{Amsterdam, Netherlands}
\acmBooktitle{Fifteenth ACM Conference on Recommender Systems (RecSys '21), September 27-October 1, 2021, Amsterdam, Netherlands}
\acmPrice{15.00}
\acmDOI{10.1145/3460231.3474254}
\acmISBN{978-1-4503-8458-2/21/09}

\graphicspath{{figs/}}

\usepackage{graphicx}
\usepackage{amsmath}
\usepackage{multirow}
\usepackage[caption=false]{subfig}
\newcommand{\donotdisplay}[1]{}
\newlength{\lgCase}

\begin{document}

\title[Information Interactions in Outcome Prediction]{Information Interactions in Outcome Prediction: Quantification and Interpretation using Stochastic Block Models}

\author{Ga\"el Poux-M\'edard}
\affiliation{%
  \institution{Université de Lyon, Lyon 2, UR 3083}
  \streetaddress{5 avenue Pierre Mendès France}
  \city{Bron}
  \country{France}}
\email{gael.poux-medard@univ-lyon2.fr}

\author{Julien Velcin}
\affiliation{%
  \institution{Université de Lyon, Lyon 2, UR 3083}
  \streetaddress{5 avenue Pierre Mendès France}
  \city{Bron}
  \country{France}}
\email{julien.velcin@univ-lyon2.fr}

\author{Sabine Loudcher}
\affiliation{%
  \institution{Université de Lyon, Lyon 2, UR 3083}
  \streetaddress{5 avenue Pierre Mendès France}
  \city{Bron}
  \country{France}}
\email{sabine.loudcher@univ-lyon2.fr}

\renewcommand{\shortauthors}{Poux-M\'edard et al.}

\begin{abstract}
In most real-world applications, it is seldom the case that a result appears independently from an environment. In social networks, users’ behavior results from the people they interact with, news in their feed, or trending topics. In natural language, the meaning of phrases emerges from the combination of words. In general medicine, a diagnosis is established on the basis of the interaction of symptoms. Here, we propose the Interacting Mixed Membership Stochastic Block Model (IMMSBM), which investigates the role of interactions between entities (hashtags, words, memes, etc.) and quantifies their importance within the aforementioned corpora. We find that in inference tasks, taking them into account leads to average relative changes with respect to non-interacting models of up to 150\% in the probability of an outcome and greatly improves the predictions performances. Furthermore, their role greatly improves the predictive power of the model. Our findings suggest that neglecting interactions when modeling real-world phenomena might lead to incorrect conclusions being drawn.
\end{abstract}



\maketitle

\section{Introduction}
With the Internet, people have begun to interact with each other as never before. Nowadays, social networks such as Facebook, Reddit or WhatsApp let us share and compare ideas. Modeling the dynamics of these exchanges can help us understanding why and how various pieces of information (e.g., hashtags, memes, ideas, etc.) will flow through a community. We refer to these pieces of information as \textit{entities}. Understanding the underlying dynamics at work provides powerful means to predict and control entities’ spread. Up to now, little work has been done on investigating the role of interacting entities in users' decisions (retweet, share, comment answer, etc.), or more generally in the probability of outcomes. A number of previous works on information diffusion theory only consider a user acting on an isolated piece of information \cite{PastorSatorras2015,pouxmdard2019influential}. On some occasions, theoretical frameworks have been developed to investigate how the presence of concurrent pieces of information affect the action a user exerts on them in a network \cite{InteractingViruses}. However, a fundamental question to be answered is how pieces of information interact in the informational landscape. Considering a recommender system, a customer that bought a smartphone might be interested by side accessories such as headphones or selfies sticks, but a customer that bought a smartphone and a camera lens extension might be more interested in buying a professional camera. 
Another example considering a disease-symptoms corpus, the diagnosis of a disease is established from the interaction of several symptoms; the interaction between ``running nose'' and ``headache'' symptoms is likely to lead to a ``flu'' diagnosis. An approach without interactions would be less successful here, since each of the products/symptoms can lead to a number of different recommendations/diagnoses. The same line of reasoning can be applied to the prediction of retweets (user exposures to A and B affects its retweet of C), music playlist building (same as before), detection of controversial posts (what combinations of words trigger what answers), etc.
Beyond the state-of-the-art guess that a single piece of information is enough to trigger another one (in music recommendation for instance, a user might associate ``ACDC'' $\rightarrow$ ``Metallica'', or ``Chopin'' $\rightarrow$ ``Mozart'') \cite{CoC}, we believe that the world is more complex, and that such a mechanism should not be so simplistic. Our idea in the present work is to take the interaction of pairs of entities into account in the prediction of an outcome (``ACDC'' + ``Metallica'' $\rightarrow$ ``Pink Floyd''). Therefore, throughout the present article we develop a novel prediction method with multiple entity types.

The remainder of this article is organized as follows. In Section~\ref{SotA}, we present an overview of the landscape of information interaction studies for both information spread and recommender systems.
Then in Section~\ref{modelDscr}, we detail our model and its mathematical derivation and describe an efficient algorithm to apply it. 
Next, in Section~\ref{Experiments}, we assess the model on real-world datasets. 
Finally, in Section~\ref{Discussion}, we quantify the role of interactions in each corpus and provide insight to analyzing the results from a semantic point of view.

\section{Background}
\label{SotA}
\subsection{Interaction in information diffusion}
Much previous research was aimed at understanding the mechanisms of information propagation. Many studies focus on the importance of the network structure coupled with the intrinsic contagiousness, or \textit{virality}, of the spreading entity (that depends on its nature and content \cite{OpinionsCoevol,SpreadNewsOnlineWWW}) in the modeling of the spreading processes \cite{DKempeMaxSpreadInfSocNet,InfoPath}. 

Recent works proposed to model the diffusion of information as the result of an interaction between information \cite{InteractingViruses,Poux2021InterRate}. Following a similar idea, Myers and Leskovec \cite{CoC} investigated interactions between contagions on Twitter. The authors aim to find the interaction factors between different tweets in one’s activity feed. Their findings suggest that interactions between tweets play a determinant role in their retweet. In their work, they assume that there is an inherent virality for every tweet (that is an inherent probability to be retweeted) computed from the frequency of retweets, to which is added a small interaction term. 

While this latter work opens paths for studying interactions, it also presents various limitations that we address here.
Firstly, the method proposed by the authors makes predictions solely based on tweets that have been observed in the feed of a given user. It therefore limits the application range of the model uniquely to systems based on the retweets (or share) concept, where information has to appear first in order to be spread. This model is hardly applicable to systems that are based on exogenous reactions (e.g., online forums, playlist building and recommender systems) where information can appear as a consequence of different entities (``Capital'' + ``Netherlands'' $\rightarrow$ ``Amsterdam''). We address this problem by developing a more global framework allowing for outputs that are different from the inputs in the considered datasets.
Secondly, interaction is defined as a correction to the frequency of retweets of a given tweet in any context. We argue this can lead to false conclusions about interactions. Imagine that interactions lead to a shift of $\Delta p$ on the base probability $p$ (virality) of a retweet, and that this interaction happens in a fraction $f$ of all observations of a given tweet being retweeted. The virality as defined in \cite{CoC} then equals $p(1-f) + (p+\Delta p) f = p + f \Delta p$, which is by construct larger than the actual probability of a retweet in the absence of interaction. Therefore, defining interaction according to this quantity is wrong (for instance it results in adding an interaction term $f \Delta p$ when there is in fact no interaction). Virality needs to be inferred by the model at the same time as the contribution of interactions to be properly defined, which is part of what we propose to do here.

\subsection{A recommender system approach}

Research in recommender systems applied to multiple pieces of information is motivated by numerous descriptive studies on multimodal networks structure \cite{YizhouHan2012,Chuan2016,Rashed2020,Huan2016}. Typically in \cite{YizhouHan2012}, the authors study interaction between multiple entity types \textit{via} a heterogeneous network representation and define clusters of entities based on structural properties of the resulting graph. 
However, as pointed out by the authors, this method is heavily influenced by the structural clustering method used --in this case a meta-path-based clustering \cite{Yizhou2011}. Moreover, defining edge weights in heterogeneous graphs is subjective and requires additional learning algorithms.

A more direct representation of real-world systems is based on collaborative filtering techniques, that directly mines clusters from pieces of information interaction patterns and generates a weighted interaction graph between entities from unit independent observations.
Typically, a widely used method in commercial applications is based on a Matrix Factorization approach \cite{MFReco}. This method considers a large number of user-item pairs and identifies regularities to model them in a lower dimensional space (e.g., it groups regularities into clusters). It has been shown that this algorithm is in fact a particular case of a wider model family: the Mixed Membership Stochastic Block Model (MMSBM) \cite{AiroldiMMSBM,HierarchicalMMSBM}. 
Another particular case of this family is the single membership Stochastic Block Model (SBM) \cite{Yuchung1987}, whose use in the discovery of underlying interaction dynamics has been suggested in recent years \cite{SEESdrugdrugInter}.

More recently, in \cite{AntoniaScalableRS} the authors develop an extension of the MMSBM considering bipartite graphs. This formulation generalizes MF models into a more global framework based on the MMSBM. In this approach, entities of different type (e.g. users and items in the case of online shopping recommendation) are grouped into distinct clusters whose interaction result in an outcome (e.g. buy or not buy). This model is optimized \textit{via} a scalable EM algorithm; it outperforms state-of-the-art models such as SBM \cite{Yuchung1987,SEESdrugdrugInter}, MF \cite{MFReco} and Mixed-Membership MF \cite{MMMF} both in performing predictions and in scalability. 

However, none of the cited works consider the interaction between pieces of information of the same nature in the prediction of an outcome. A drug might interact with another one, but the joint interaction of two drugs on a third one cannot be investigated \cite{SEESdrugdrugInter}. A user on Netflix is predicted to like a given movie because she is partially part of the group that liked the movie A and partially part of the group that disliked movie B, but all the user groups are independent from one another \cite{MFReco,AntoniaScalableRS}. Friendship between individuals is determined on the basis of the independent groups of friends they belong to, but not on the basis of the joint belonging to various groups \cite{AiroldiMMSBM,HierarchicalMMSBM,Jamali2011}. 
Typically in \cite{AntoniaScalableRS}, embedding pieces of information of same nature in a bipartite graph seems irrelevant: taking the example of a online purchase recommendation system, a product should not interact differently with an other (or belong to different clusters) because it is on the left or the right side of a bipartite graph. Instead, we want to enforce a symmetrical interaction and thus a single clustering for pieces of information that interact in this way.

Relaxing the independence assumption and looking at the probability of an outcome due to the \textit{joint} membership to more than one group (while also allowing for outcomes resulting that do not result from any interaction) provides a better assumption for modeling the subtle interaction process. Taking back the previous example of medical diagnosis: a disease can seldom be diagnosed on the basis of a single symptom (e.g. without interaction between symptoms), but rather on a symmetric combination of symptoms (e.g. with interaction). 

\subsection{Contributions}
\begin{itemize}
\item We develop a scalable model that accounts for interactions between entities: the \textbf{I}nteracting \textbf{M}ixed \textbf{M}embership \textbf{S}tochastic \textbf{B}lock \textbf{M}odel (IMMSBM).
\item We show by comparing our results with non-interacting MMSBM \cite{AntoniaScalableRS} that taking interaction into account leads to a more accurate recommendations in real-world situations on 4 different datasets.
\item We provide a proper way to infer virality, which allows to correctly define and compute interaction terms.
\item We quantify the role of interactions in various corpus and highlight the necessity of taking them into account when dealing with real-world spreading or recommendation problems.
\end{itemize}

\section{The IMMSBM model}
\label{modelDscr}
In this section, we develop the IMMSBM model. We propose an approach based on standard Mixed Membership Stochastic Block Modeling \cite{AiroldiMMSBM,AntoniaScalableRS}, which we modify in order to take interactions into account. Building our model on the MMSBM allows to assume that each entity does not have only one membership, which is in line with the real situation.
The IMMSBM requires no prior information on the system and its numerical implementation is possible via a scalable Expectation-Maximization algorithm of linear complexity with the size of the dataset. In addition to the state-of-the-art problems which our method answers, it also offers better predictive power than non-interacting baselines.

\begin{figure}
\centering
    \subfloat{
        \includegraphics[width=.45\textwidth]{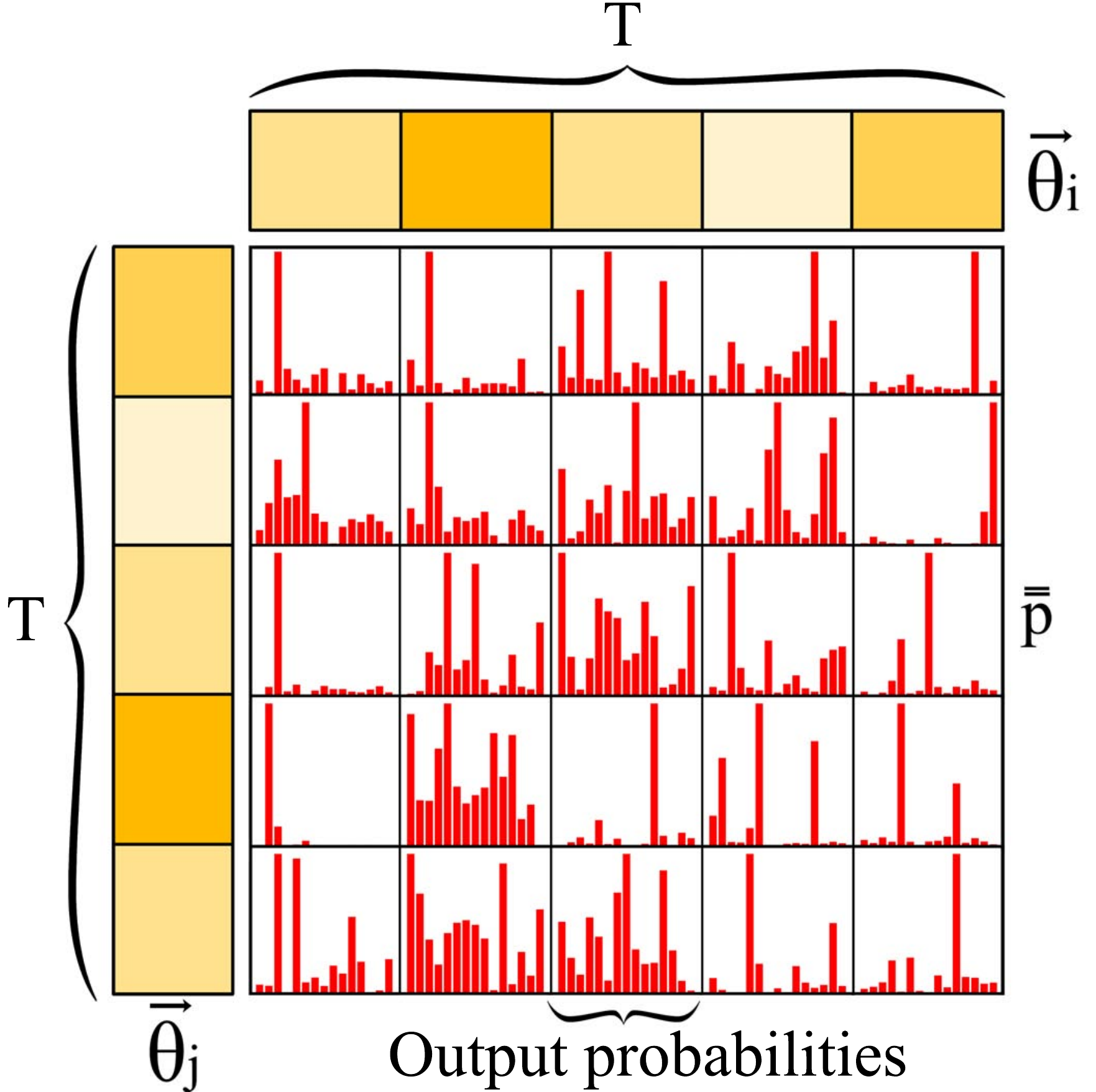}
    }
    
    \subfloat{
        \includegraphics[width=.37\textwidth]{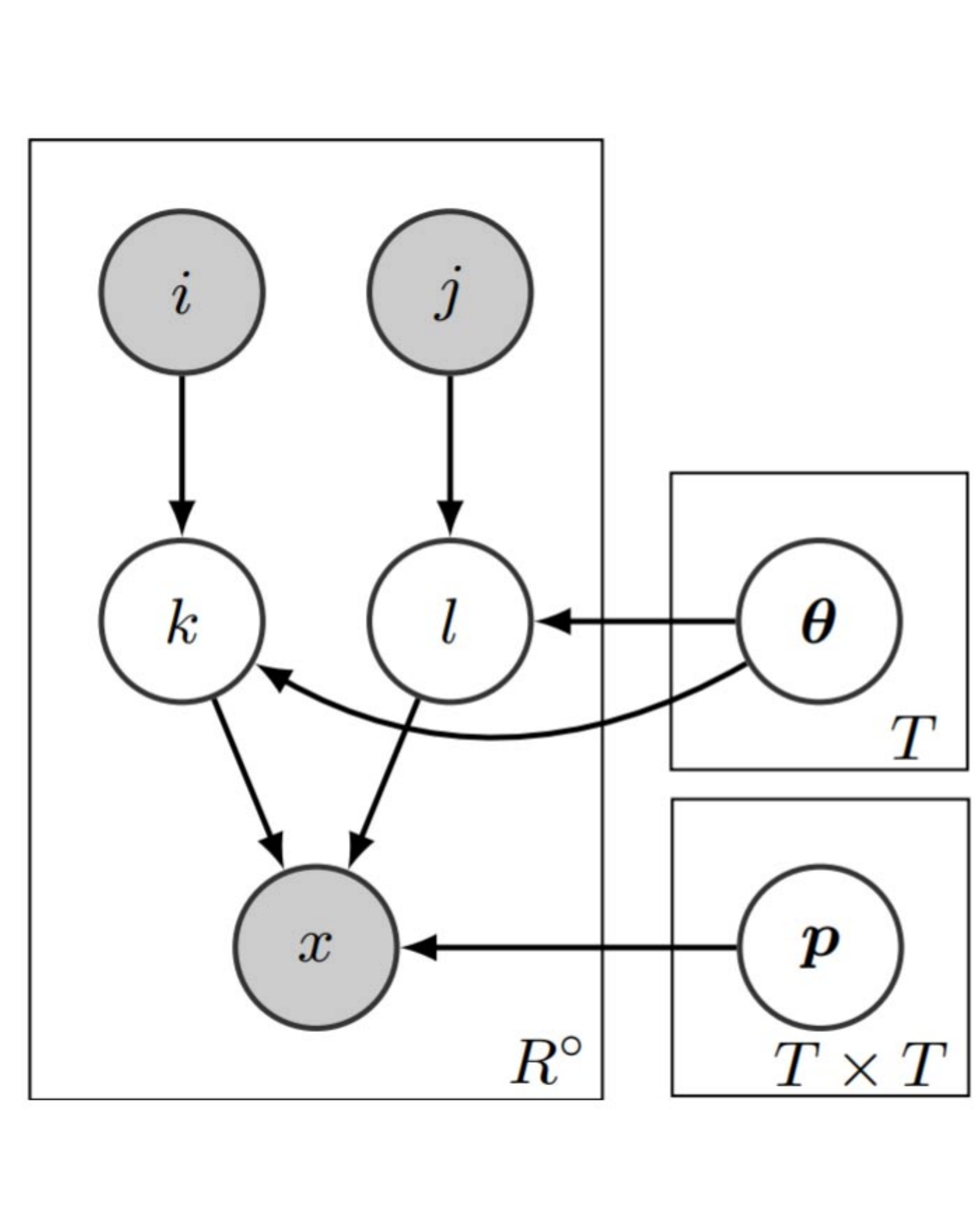}
    }
    \caption{\textbf{Illustration of the model} -- \textit{(Top)} Schema of IMMSBM for a single pair of entities $(i,j)$ (which could be ``fever'' and ``cough'' for instance). Input entities are grouped into T clusters in different proportions; the proportion to which they belong to each cluster is quantified by a $\theta$ matrix (dimension $[I \times T]$ where $I$ is the input space). The clusters then interact to generate a probability distribution over the output entities defined by the interactions tensor \textbf{p} (dimension $[T \times T \times O]$ where $O$ is the output space). \textit{(Bottom)} Alternative representation of the IMMSBM as a graphical model. To generate each output, for each observation ($i$, $j$, $x$) in the set $R^{\circ}$, a cluster ($k$ and $l$) is drawn for each input entity ($i$, $j$) from a distribution encoded in the matrix $\theta$. The generated output $x$ is drawn from a multinomial distribution conditioned by the previously drawn clusters k and l encoded in $\Vec{p}$.} \label{figModel}
\end{figure}

The goal of the model is to predict the most likely result of an interaction between two entities (i and j in Fig.\ref{figModel}). In recommender systems, it would recommend the song the most likely to fit user's tastes given previous songs they listened to, or the product the most likely to be bought given a user's purchase history. Another example about assisted medical diagnosis: observing the words ``fatigue'' and ``cough'' in a medical report is more likely to imply the observation of ``flu'' than ``anemia'', despite ``anemia'' being often associated with the ``fatigue'' symptom. The model will group data into clusters (membership matrix $\theta$ Fig.\ref{figModel}) that interact symmetrically with each other (interactions tensor \textbf{p}), resulting in a probability over the possible outputs to appear (histograms Fig.\ref{figModel}-top). We have no prior knowledge of the content of the groups, and we only need to set the number of clusters T.
\label{secDerivEM}

\subsection{Interacting MMSBM}
\label{likelihoodSec}

We refer to the interacting entities as input entities $(i,j) \in I^2$, and aim to predict an output entity $x \in O$. $I$ is the input space (the entities that interact: products, symptoms or songs for instance) and $O$ is the output space (the entities resulting from the interaction in an answer, diagnosed diseases for instance). We illustrate in Figure~\ref{figModel}-top how the input space and output space are related according to our model: input entities are clustered, and the interaction between those clusters give rise to probability distributions over the output entities. Note that $I$ and $O$ can be identical or radically different according to what we want to model. In the case of musical recommendation, $I$ accounts for observed artists in a user's feed, and $O$ for recommended artists --that do not necessarily appears in the user's feed, unlike \cite{CoC}. In the case of medical diagnosis, $I$ accounts for symptoms (fever, cough, anemia, etc.) and $O$ for diseases (Alzheimer, hepatitis, etc.). 
If one wants to predict named entities in the answer to a Reddit post based on interacting named entities in the original post, then $I=O$ since the predicted vocabulary can also be used as an input (e.g., the named entity ``Sun'' can either be an input that interacts with other inputs, or an output). As an alternative visualization, we present the graphical generative model of the IMMSBM in Figure~\ref{figModel}-bottom. The observed data then takes the form of triplets ($i$, $j$, $x$) signifying that the combined presence of input entities i and j leads to the output entity x.
A given triplet can obviously occur several times in the same dataset.

We assume there are regularities in the studied dataset, so that given inputs interact with each other in the same way. Their classification into clusters would therefore be relevant. For the medical dataset example, this means that symptoms such as ``fever'' and ``pallor'' often come in pairs and therefore are considered as similar regarding the diagnosis; they would belong to the same cluster. We define the membership matrix $\theta$ associating each input entity with clusters in different proportions, such that $\theta_i$ is a $[1 \times T]$ vector with $\sum_t^T \theta_{i,t}=1$. Note that unlike in single membership stochastic block models an entity does not have to belong to only one group \cite{ReviewSBM}. Given the possible semantic variation of entities (polysemy of words in natural languages -e.g. ``like'', ``swallow''-, symptoms with various causes in medicine -``headache'', ``fever''-, etc.), an approach \textit{via} a mixed-membership clustering seems more relevant.

We then define the cluster interactions tensor $p_{k,l}(X_{k,l}=x)$ of dimensions $[T \times T \times O]$ as the probability that the interaction between clusters $k \in T$ and $l \in T$ gives rise to the output $x \in O$.
A property of this definition is that $\sum_{x} p_{k,l}(X_{k,l}=x) = 1 \ \forall k,l$. The role of those two quantities is schematized in Fig.\ref{figModel}.

We choose to consider only one membership matrix $\theta$ for all of the inputs, instead of one per input entry as in \cite{AntoniaScalableRS}. It implies the assumption that interactions to be symmetric, which means that an observation ($i$, $j$, $x$) is equivalent to ($j$, $i$, $x$). This follows the idea of \cite{InteractingViruses} where it is assumed that the interaction between two viruses is symmetric, meaning the interaction influences both viruses with the same magnitude. Therefore, there is no need to consider a different clustering for inputs i and j, which motivates the use of a single membership matrix $\theta$.

We now propose to define the entities interactions tensor $P_{i,j}(X_{i,j}=x)$, representing the probability that the interaction between inputs i and j implies the output x as:
\begin{equation}
    P_{i,j}(X_{i,j}=x) = \sum_{k,l} \theta_{i,k}\theta_{j,l}p_{k,l}(X_{k,l}=x)
\end{equation}
For the sake of brevity, from now on we will refer to $p_{k,l}(X_{k,l}=x)$ as $p_{k,l}(x)$. We define the likelihood of the observations given the parameters as:
\begin{equation}
    \label{eqLikelihood}
    P(R^{\circ} \vert \theta, p) = \prod_{(i,j,x) \in R^{\circ}} \sum_{k,l} \theta_{i,k}\theta_{j,l}p_{k,l}(x)
\end{equation}
where $R^{\circ}$ denotes the set of triplets in the training set (input, input, output). Note that the remaining triplets $R \setminus R^{\circ}$ are used as test set.

Importantly, this formulation differs from related work in that the model presented here considers a symmetric interaction between pieces of information in the prediction of an outcome instead of their asymmetric combination \cite{AntoniaScalableRS}. Considering a musics recommender system for instance, clustering the same songs using two independent membership matrices (which is the formulation introduced in \cite{AntoniaScalableRS}) would make few sense and might eventually lead to incorrect results. Indeed, since the convergence towards a global optimum is not guaranteed, such algorithm might assign different memberships to a song depending on whether it is on the left or right side of a triplet. Put differently, we would have different output probabilities for $(i,j,x)$ and $(j,i,x)$, which is something we avoid here. In our formulation, only the outputs \textit{can} be of different nature than the inputs.

\subsection{Inference of the parameters}
We are now looking at inferring the model parameters, namely the components of the interactions tensor $ P_{i,j}(X_{i,j}=x)$. As stated in the previous section, we decomposed this tensor into an algebraic combination of the matrix $\theta$ and the tensor $p$. In this section, we introduce a method to infer them via an Expectation-Maximization (EM) algorithm. EM is a 2-step iterative algorithm. The first step consists of computing the expectation of the likelihood with respect to the latent variables, denoted $\omega_{i,j,x}(k,l)$, with parameters $\theta$, p set as constant. The second step consists of maximizing this expectation of the likelihood with respect to the parameters $\theta$, p. Iterating this process guarantees that the likelihood converges towards a local maximum.

\subsubsection{Expectation step}
Taking the logarithm of the likelihood as defined in Eq.\ref{eqLikelihood}, denoted $\ell$, we have:
\begin{equation}
\label{eqJensen}
    \begin{split}
         \ell &= \sum_{(i,j,x) \in R^{\circ}} \ln \sum_{k,l} \theta_{i,k}\theta_{j,l}p_{k,l}(x) \\
         &= \sum_{(i,j,x) \in R^{\circ}} \ln \sum_{k,l} \omega_{i,j,x}(k,l) \frac{\theta_{i,k}\theta_{j,l}p_{k,l}(x)}{\omega_{i,j,x}(k,l)} \\
         & \geq \sum_{(i,j,x) \in R^{\circ}} \sum_{k,l} \omega_{i,j,x}(k,l) \ln \frac{\theta_{i,k}\theta_{j,l}p_{k,l}(x)}{\omega_{i,j,x}(k,l)}
    \end{split}
\end{equation}
We used Jensen's inequality to go from the 2nd to 3rd line.
The inequality in Eq.\ref{eqJensen} becomes an equality for: \begin{equation}
\label{eqOmega}
    \omega_{i,j,x}(k,l) = \frac{\theta_{i,k}\theta_{j,l}p_{k,l}(x)}{\sum_{k',l'} \theta_{i,k'}\theta_{j,k'}p_{k', l'}(x)}
\end{equation}
where $\omega_{i,j,x}(k,l)$ is interpreted as the probability that the observation $(i,j,x)$ is due to i belonging to the group $k$ and j to $l$, that is the expectation of the likelihood of the observation $(i,j,x)$ with respect to the latent variables $k$ and $l$. Therefore, Eq.\ref{eqOmega} is the formula for the expectation step of the EM algorithm. An alternative derivation of the expectation step formula is discussed in \cite{Bishop}.

\subsubsection{Maximization step}
\label{maximization}
This step consists in maximizing the likelihood using the parameters of the model $\theta$ and p, independently of the latent variables.
In order to take into account the normalization constraints, we introduce the Lagrange multipliers $\phi$ and $\psi$. Following this, the constrained log-likelihood reads:
\begin{equation}
\label{eqLagMult}
    \ell_c = \ell - \sum_i (\phi_i \sum_{t} \theta_{i,t} - 1) - \sum_{k,l} (\psi_{k,l} \sum_{x} p_{k,l}(x) - 1)
\end{equation}
We first maximize $\ell_c$ with respect to each entry $\theta_{mn}$.
\begin{equation}
\label{eqDerivTheta}
    \begin{split}
        \frac{\partial \ell_c}{\partial \theta_{mn}} =& \frac{\partial \ell}{\partial \theta_{mn}} - \phi_m = 0\\
        =& \sum_{\partial m} \left( \sum_{l} \frac{\omega_{m,j,x}(n,l)}{\theta_{mn}} + \sum_{k} \frac{\omega_{j,m,x}(k,n)}{\theta_{mn}} \right) - \phi_m\\
        =& \frac{1}{\theta_{mn}}\sum_{\partial m} \left( \sum_{t} \omega_{m,j,x}(n,t) + \omega_{j,m,x}(t,n) \right) - \phi_m\\
        \Leftrightarrow\ \theta_{mn} =& \frac{1}{\phi_m}\sum_{\partial m} \sum_{t} \omega_{m,j,x}(n,t) + \omega_{j,m,x}(t,n)
    \end{split}
\end{equation}
where $\partial m$ stands for the set of observations in which the entry m appears $\{ (j,x) \vert (m,j,x) \in R^{\circ} \}$. Note that summing over this set in line 2 of Eq.\ref{eqDerivTheta} implies that the relation between two inputs entities is symmetric --if $(i,j,x) \in R^{\circ} \ \text{ implies } (j,i,x) \in R^{\circ}$. Summing the last line of Eq.\ref{eqDerivTheta} over $n \in T$ then multiplying it by $\phi_m$, we get an expression for $\phi_m$:
\begin{equation}
        \phi_m\sum_n \theta_{mn} = \phi_m = \sum_{\partial m} \sum_{t,n} \omega_{m,j,x}(n,t) + \omega_{i,m,x}(t,n) = \sum_{\partial m} 2 = 2 \cdot n_m
\end{equation}
Where $n_m$ is the total number of times that m appears as an input in $R^{\circ}$. Finally, plugging back this result in Eq.\ref{eqDerivTheta}, we get:
\begin{equation}
    \label{eqMaxTheta}
    \theta_{mn} = \frac{\sum_{\partial m} \left( \sum_{t} \omega_{m,j,x}(n,t) + \omega_{i,m,x}(t,n) \right)}{2.n_m}
\end{equation}

Following the same line of reasoning for p, we get:
\begin{equation}
    \label{eqMaxp}
    p_{r,s}(l) = \frac{\sum_{\partial l} \omega_{i,j,l}(r,s)}{\sum_{(i,j,l') \in R^{\circ}} \omega_{i,j,l'}(r,s)}
\end{equation}
The set of Equations \ref{eqMaxTheta} and \ref{eqMaxp} constitutes the maximization step of the EM algorithm. They hold only if the input entities interactions are symmetric (e.g. when $\{ (j,x) \vert (m,j,x) \in R^{\circ} \} = \{ (i,x) \vert (i,m,x) \in R^{\circ} \}$), which is what we aimed to do.
It is worth noting that the proposed algorithm offers linear complexity with the size of the dataset $\mathcal{O}(\vert R^{\circ} \vert)$ provided the number of clusters is constant, while guaranteeing convergence to a local maximum.

\begin{figure*}[h]
    \includegraphics[width=1.0\textwidth]{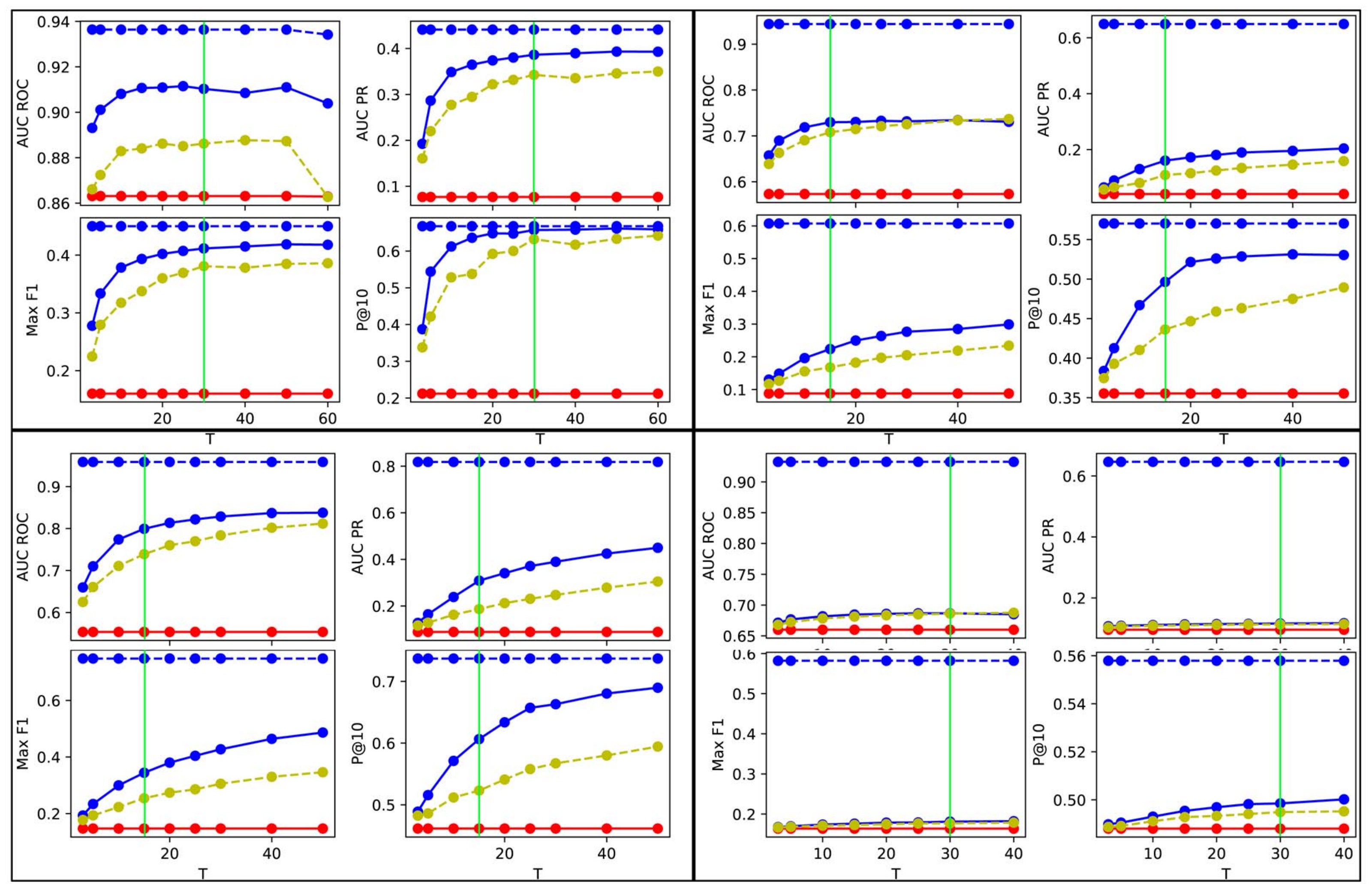}
    \caption{Performance variations on all the metrics for every dataset considered. Dashed blue line: upper limit to performances ; blue line: IMMSBM ; yellow dashed line: MMSBM ; red line: naive baseline. Top left: PubMed ; top right: Spotify ; bottom left: Twitter ; bottom right: Reddit. The vertical green line shows the selected number of clusters; it is chosen using the AIC criterion, which matches with the elbow of the various metrics considered.} \label{figCalib}
\end{figure*}
\donotdisplay{
\subsection{Expectation-Maximization algorithm}
To derive EM equations, we use a standard variational approach. Using Jensen's inequality and Lagrange multipliers in a similar fashion as in \cite{Bishop} and \cite{AntoniaScalableRS}, we get the set of equations \ref{EMEq}. Complete derivation is done in SI, Section~1.

\begin{equation}
    \label{EMEq}
    \begin{cases}
        \omega_{i,j,x}(k,l) &= \frac{\theta_{i,k}\theta_{j,l}p_{k,l}(x)}{\sum_{k',l'} \theta_{i,k'}\theta_{j,k'}p_{k', l'}(x)} \text{ (E-step)}\\
        \theta_{mn} &= \frac{\sum_{\partial m} \left( \sum_{t} \omega_{m,j,x}(n,t) + \omega_{i,m,x}(t,n) \right)}{2.n_m} \text{ (M-step)}\\
        p_{r,s}(l) &= \frac{\sum_{\partial l} \omega_{i,j,l}(r,s)}{\sum_{(i,j,l') \in R^{\circ}} \omega_{i,j,l'}(r,s)}\text{ (M-step)}
    \end{cases}
\end{equation}
}

\section{Experiments}
\label{Experiments}
\subsection{Datasets and evaluation protocol}
We test the performance of our model on 4 different datasets. The first dataset (\textbf{PubMed} dataset) is built with 15,809,271 medical reports collected from the PubMed database as a good approximation for human-disease network \cite{SymptDiseNet}. This dataset is not explicitly about recommender systems, but provides an intuitive way to understand how our recommendation approach works by suggesting likely diseases given a collection of symptoms. 
The second (\textbf{Twitter} dataset) with 139,098 retweets gathered in october 2010 associated with the 3 last tweets in the feed preceding each retweet \cite{DataSetTwitter}. The task is to infer tweets a user is the most likely to retweet. A possible application would be a personnalized recommendation of such tweets in the ``Trends for you'' Twitter section. 
The third (\textbf{Reddit} dataset) with the entirety of posts in the subreddit r/news in May 2019 (225,485 message-answer relationships in total). We aim at predicting the content of the answer given the incoming message. A possible application would be similar as what Gmail does when suggesting automated answers to an email given keywords present in the message.
Finally, the last (\textbf{Spotify} dataset) is built with 2,000 music playlists associated with keywords ``english'' and ``rock'' of random Spotify users. We predict the next song a user will add to a playlist given this user's history. A RS application would suggest a ranked list of songs to the user is likely to add to a playlist.
Each dataset is formed by associating every pair of inputs in a \textit{message} (i.e. a list of symptoms, a user's feed, a Reddit post, and a playlist's last artists) with an \textit{answer} (i.e. a disease, a retweet, a Reddit answer, an artist added to a playlist). The building process of datasets is further described in SI, Section~1, together with direct links to access them for possible replication studies \footnote{To ease eventual replication studies, we provide the Python implementation of the IMMSBM and of the baseline model used in this article in a Git repository at https://github.com/GaelPouxMedard/IMMSBM.}. 

From the raw datasets, we form the test set by randomly sampling 10\% of the coupled structures (message$\rightarrow$answer). The 90\% coupled structures left are used as a training set. The number of clusters is determined using the elbow method. For each corpus: 30 for PubMed, 15 for Twitter, 30 for Reddit, and 15 for Spotify. We perform 100 independent runs, each with independent random initialization of the parameters $\theta$ and $p$.  The EM loop stops once the relative variation of the likelihood between two iterations is less than 0.001\%.

\subsection{Baselines}
\label{baselines}
\subsubsection{Naive baseline}
The naive baseline is simply the frequency of the outputs in the test set. This naive classifier predicts the value of every output independently of the inputs.

\subsubsection{MMSBM} 
We use the classical MMSBM as a second baseline. In this formulation interactions are not taken into account and is identical to state-of-the-art work done in recommender systems \cite{AiroldiMMSBM,HierarchicalMMSBM,AntoniaScalableRS}. Instead of considering triplets (input, input, output), we instead here apply the classical MMSBM on pairs (input, output). This baseline is exactly the same as the model introduced in \cite{AiroldiMMSBM}. We then train our baseline whose log-likelihood is defined as $\ell_{BL} = \sum_{(i,x) \in R^{\circ}} \ln \sum_{k} \theta_{ik} p_k(x)$ on the same datasets as in the main experiments. We infer the parameters \textit{via} an Expectation-Maximization algorithm similar to the one described in Section~\ref{secDerivEM}. The number of clusters being the only parameter to be tuned. It is chosen in the same way as it is for the IMMSBM --by minimizing the AIC criterion \cite{Akaike1973}. We see further in the experiments that this criterion matches with the elbow of the various metrics. We make 100 independent runs with random initialization and discuss here the results of the highest-likelihood run. This baseline provides an alternative way to quantify the importance of interactions by comparison with the case where these are taken into account. We expect the baseline model to find results that are equivalent to the diagonal probabilities of the main model (i.e. similar to $P_{i,i}(x) = \sum_{k,l}\theta_{i,k}\theta_{i,l}p_{k,l}(x)$). This is because the diagonal $P_{i,i}(x)$ is supposed to account for the apparition of x given only the presence of i, which is what this baseline computes. Furthermore, this model will provide insight to the generalization of the assumption made in Myers and Leskovec \cite{CoC}, stating that the probability of an output is greatly dominated by a prior quantity unrelated to any interaction (known as \textit{virality}, see Section~\ref{SotA}).

\subsubsection{Perfect modeling - Upper limit to prediction}
We also compare our results to a mathematical upper limit to our predictions. While in some systems the predictions could theoretically be perfect, in most situations the dataset simply does not allow it. Consider as an example a case where the test set contains twice the triplet (``fever'', ``pallor'', ``influenza'') and once the triplet (``fever'', ``pallor'', ``anemia''): the model cannot make a prediction better than 66\%. In SI, Section~2, we develop a general method to derive such an upper limit to predictions for a given dataset. This upper limit is mathematically the best performance any model can do given a dataset structure.

\subsection{Results}
The metrics we use to assess the performance of our model are the max-F1 score, the area under the receiver operating characteristic curve (ROC) curve and the precision@10. For all of these quantities, the closer to 1 they are, the better the performance.

For evaluation, we adopt a guessing process as follows. For every pair of inputs, we compute the probability vector for the presence of every possible output. Then we predict all of the probabilities larger than a given threshold to be ``present'', and all the others to be ``absent''. Comparing those predictions with the observations in the test set, we get the confusion matrix for the given threshold. We then lower the threshold and repeat the process to compute the various metrics. 

We recall that we do not compare our approach to \cite{CoC} because its formulation does not allow to make prediction on exogenous output, that is when the output is not part of the input pair. The more general model presented here is needed in order to consider the effect of interactions on the probability of exogenous outputs.

\begin{table*}
\caption{Experimental results for the four metrics considered, from each model applied to each corpus. We see that our model outperforms the proposed baselines in every dataset for almost every evaluation metric -- the error bars overlap for the AUC (ROC) on the PubMed corpus. The given error corresponds to the standard deviation over the 10 runs. The naive baseline and upper limit results are constant over the runs and therefore have no variance. \label{tabMetrics}}
\centering
\setlength{\lgCase}{3.25cm}
\begin{tabular}{ |p{0.08\lgCase}|p{0.6\lgCase}|p{\lgCase}|p{\lgCase}|p{\lgCase}|p{0\lgCase}}
\cline{3-5}
  \multicolumn{0}{c}{\rotatebox[origin=c]{90}{}} & \multicolumn{0}{c|}{\rotatebox[origin=c]{90}{}} & \centering P@10 & \centering Max-F1 & \centering AUC (ROC) & \\
 \cline{1-5}
    \centering\multirow{4}{*}{\centering\rotatebox[origin=c]{90}{\textbf{PubMed}}} & \centering Naive & \centering 0.212 & \centering 0.160 & \centering 0.863 & \\
     
     & \centering MMSBM & \centering 0.627 $\pm$ 0.002 & \centering 0.393 $\pm$ 0.002 & \centering \textbf{0.911 $\pm$ 0.000} & \\
     
     & \centering IMMSBM & \centering \textbf{0.656 $\pm$ 0.001} & \centering \textbf{0.411 $\pm$ 0.001} & \centering \textbf{0.911 $\pm$ 0.002} & \\
     
     & \centering Up.lim. & \centering 0.668 & \centering 0.450 & \centering 0.936 & \\
 \cline{1-5}

     \centering\multirow{4}{*}{\centering\rotatebox[origin=c]{90}{\textbf{Twitter}}} & \centering Naive & \centering 0.462 & \centering 0.147 & \centering 0.554 & \\
    
     & \centering MMSBM & \centering 0.529 $\pm$ 0.005 & \centering 0.254 $\pm$ 0.005 & \centering 0.741 $\pm$ 0.004 & \\
     
     & \centering IMMSBM & \centering \textbf{0.610 $\pm$ 0.004} & \centering \textbf{0.349 $\pm$ 0.006} & \centering \textbf{0.800 $\pm$ 0.001} & \\
     
     & \centering Up.lim. & \centering 0.737 & \centering 0.748 & \centering 0.959 & \\
 \cline{1-5}
 
     \centering\multirow{4}{*}{\centering\rotatebox[origin=c]{90}{\textbf{Reddit}}} & \centering Naive & \centering 0.488 & \centering 0.164 & \centering 0.660 & \\
    
     & \centering MMSBM & \centering 0.495 $\pm$ 0.000 & \centering 0.177 $\pm$ 0.000 & \centering 0.686 $\pm$ 0.000 & \\
     
     & \centering IMMSBM & \centering \textbf{0.499 $\pm$ 0.000} & \centering \textbf{0.181 $\pm$ 0.000} & \centering \textbf{0.687 $\pm$ 0.000} & \\
     
     & \centering Up.lim. & \centering 0.558 & \centering 0.582 & \centering 0.933 & \\
 \cline{1-5}
 
     \centering\multirow{4}{*}{\centering\rotatebox[origin=c]{90}{\textbf{Spotify}}} & \centering Naive & \centering 0.355 & \centering 0.088 & \centering 0.573 & \\
    
     & \centering MMSBM & \centering 0.426 $\pm$ 0.006 & \centering 0.167 $\pm$ 0.003 & \centering 0.707 $\pm$ 0.002 & \\
     
     & \centering IMMSBM & \centering \textbf{0.502 $\pm$ 0.006} & \centering \textbf{0.228 $\pm$ 0.005} & \centering \textbf{0.723 $\pm$ 0.002} & \\
     
     & \centering Up.lim. & \centering 0.570 & \centering 0.607 & \centering 0.944 & \\
 \cline{1-5}

\end{tabular}

\end{table*}

In Table \ref{tabMetrics}, we show the performances of our model compared with the baselines introduced in Section~\ref{baselines}. We see that IMMSBM outperforms all of the baselines in most cases. A remarkable feature is the good P@10 of our model, primordial in many applications such as diseases diagnosis or in recommender systems. As expected, taking interactions between entities into account leads to an improved accuracy on the prediction of the missing data. This correlates to the conclusions drawn in Myers and Leskovec, stating the importance of interactions in real-world phenomenon modeling \cite{CoC}. Overall, our model performs better in many different applications for most of the corpora considered. It systematically yields a better P@10 and F1 score, making it of use in recommender systems applications. The AUC (ROC) score can be interpreted as the quality of classification of low-probability events. Overall, little improvement is observed on this aspect compared to the non-interacting baseline.

We notice however that accounting for interactions did not lead to a significant improvement in performance over the two baselines on the Reddit corpus. Given the obvious fact that language is partially formed by interacting named entities, this result may look surprising. We attribute the lack of improvement when considering interactions to the dataset provided. Indeed the dataset contains few observations for every possible pair, due to the wide range of available vocabulary of natural language \cite{CombinLanguage}. It is likely that we did not train the model with enough data for it to learn actual regularities in pair interactions. This can also be seen during the building of the test set: approximately one half of its pairs have never been observed in the training set. Therefore, we suppose that the model simply lacks enough data to identify generalities. In future work, it might be interesting to answer this problem by considering a corpus of pre-clustered entities instead of independent named entities, hence reducing the vocabulary range and adding to the regularity of the dataset.

\section{Discussion}
\label{Discussion}
\label{impInter}

\begin{figure*}
    \centering
    \includegraphics[width=1.0\textwidth]{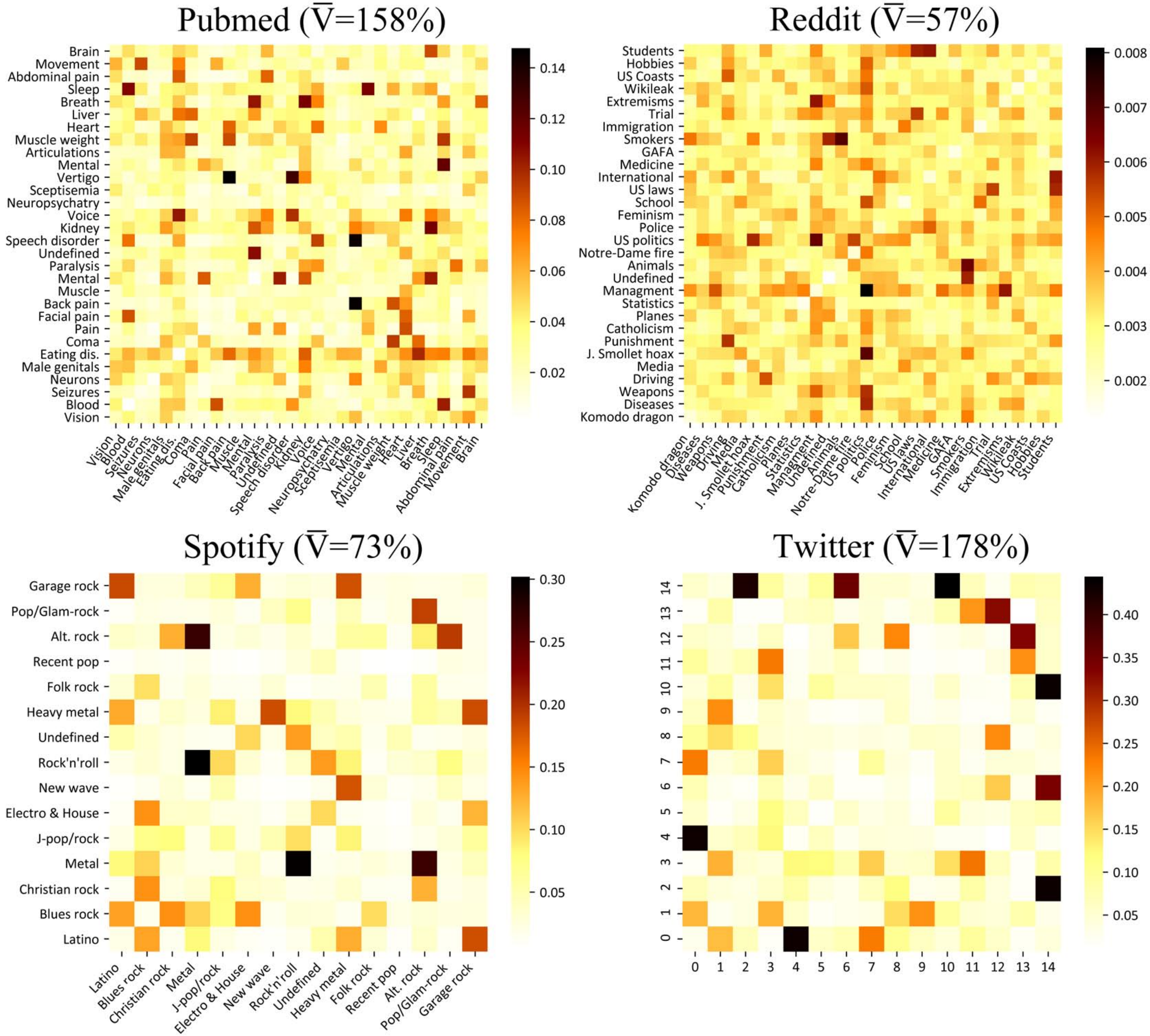}
    \caption{\textbf{Importance of interactions} - Contribution of each pair of clusters $V_{k,l}$ (heatmap) and average impact of the interactions $\bar V$ (on the right) in outcome probabilities for each corpus. Clusters typically interact with a limited number of others; these interactions still play a significant role in outcomes probabilities. The cluster have been annotated manually.}
    \label{figGpesInter}
\end{figure*}
\subsection{Relative importance of interactions}
\subsubsection{Global importance of interactions}
We recall the argument exposed in the introduction that interactions have to be inferred along with the virality of a piece of information, which cannot be calculated heuristically. Imagine that interactions lead to a shift of $\Delta p$ on the base probability $p$ (virality) of a retweet, and that this interaction happens in a fraction $f$ of all observations of a given tweet being retweeted. The virality as defined in \cite{CoC} then equals $p(1-f) + (p+\Delta p) f = p + f \Delta p$, which is by construct larger than the actual probability of a retweet in the absence of interaction. Therefore, defining interaction according to this quantity is wrong (for instance it results in adding an interaction term $f \Delta p$ when there is in fact no interaction). Virality needs to be inferred by the model at the same time as the contribution of interactions to be properly defined.

Our model infers virality along with interaction terms and yield better results than state-of-the-art methods (see Table \ref{tabMetrics}), which provides solid ground for analyzing the effect of information interaction. Close analysis of interactions between pieces of informations has been little considered in literature --what lexical fields, groups of symptoms, musical genres, kind of tweets interact with each other.
We can evaluate the importance of interactions between entities on the basis of inferred virality. We consider two quantities within each corpus: the overall relative impact of the interactions on the probability of an outcome and the contribution of each pair of groups in the modification of outcome probabilities.

To evaluate the global impact of taking the interactions into account, we compute the relative change of probability according to the virality for each triplet and average this quantity over all the triplets in the corpus. We note this quantity $\bar V$:
$$
\bar V = \frac{1}{\vert R^{\circ} \vert} \sum_{(i,j,x) \in R^{\circ}} \frac{\vert P_{i,i}(x) - P_{i,j}(x) \vert}{P_{i,i}(x)} 
$$
where $P_{i,j}(x) = \sum_{k,l}\theta_{i, k} \theta_{j, l} p_{k,l}(x)$ denotes the probability of outcome $x$ given the entities $i$ and $j$; as shown in the previous section, the diagonal elements $P_{i,i}(x)$ account for the virality of $i$ on $x$. 
The results are shown Fig.\ref{figGpesInter}.
It results that for every corpus, interaction between entities exerts a non-negligible influence on the probability of an output. Those results confirm previous work done on interactions modeling, stating the importance of taking interactions into account when analyzing real-world datasets \cite{CoC}. Interactions increase the virality of an output by a factor by 2.58 in the PubMed corpus, 2.78 in the Twitter corpus, 1.73 in the Spotify corpus and 1.57 in the Reddit corpus. Interactions have a greater effect on output probabilities for PubMed and Twitter corpora, and a lesser role for the Spotify and Reddit ones. Besides, our model applied to a dataset where interactions do not play any role ($\bar V = 0$) reduces to the non-interacting MMSBM baseline. This metric therefore allows to assert the importance of the interactions in a given corpus. 

\subsubsection{Which clusters interact}
To evaluate the role of each pair of groups plays in the modification of output probability, we consider the following quantity:
$$
V_{k,l} = \frac{\sum_{(i,j,x) \in R^{\circ}} \theta_{i,k} \theta_{j,l} (p_{k,l}(x) - P_{i,i}(x))}{\sum_{(i,j,x) \in R^{\circ}} \theta_{i,k} \theta_{j,l}}
$$
where $\theta$, $p$ and $P_{i,i}(x)$ are defined as before. This quantity is the weighted average of the change in output probability with respect to virality due the interaction between every pair of clusters $(k,l)$ to which belong entities $(i,j)$ in a proportion ($\theta_{ik}$,$\theta_{jl}$). The results are shown in the heatmaps Fig\ref{figGpesInter}. Note that the clusters have been manually given a name on the basis of their composition; labeling is subjective.

We see that most of the clusters do not interact with each other; the interactions essentially take place between a limited number of clusters. Typically, a cluster interacts significantly with only one or two other clusters in every corpus (``Vertigo'' and ``Speech disorder'' in PubMed, ``Students'' and ``Schools'' in Reddit, etc.). We also notice that in each corpus, the model forms some non-interacting cluster with low values of $V_{k,l}$ (``Neuropsychatry'' in PubMed, ``Recent pop'' in Spotify, etc.); for those, the probability of an output is essentially equal to the virality of this output. We also notice that the diagonal of the $V_{k,l}$ matrices have low values; this expresses that the interaction of a group with itself leads to an output probability close to its virality. To picture how this makes sense, we can imagine diagnosing a disease on the basis of two ear-related symptoms (``earache'' and ``hearing disorders''): the diagnosis is likely to be related to the ear as we would have guessed with only one symptom (its probability equals virality). Now imagine two symptoms of different kind (``earache'' and ``speech disorder''): the diagnosis is then likely to be related to the brain and less to the ear, so the interaction lowers the base probability (virality) of the ``ear disease'' output and increases the one of the brain disease.

Being able to see in details the extent to which interactions exert an influence in a corpus and between which categories they take place open new perspectives in research. Understanding and explaining underlying mechanisms ruling corpora is a desirable property of AI models in social sciences for instance \cite{SergioSocialDilemma,MrBanks,HumanPrefSBM}. As we demonstrated, the present work follows this line by developing an explainable model for investigating interactions.

\subsection{Entropy of membership}
Another interesting quantity to look at when considering the results is the membership entropy of the entities. This quantity measures how entities are spread over all the clusters; when this value is low, it means that the model finds strong regularities in the corpus, and the clusters are likely to be readily interpretable. Therefore, we use the normalized Shannon entropy of memberships of user i, $S_{i}^{(m)}$, whose formula is:
\begin{equation}
    S_{i}^{(m)} = \frac{1}{\log_2 \frac{1}{T}}\sum_t^T \theta_{i,t} \log_2 \theta_{i,t}
\end{equation}
Here the lowest entropy reachable is 0, which corresponds to an entity belonging to only one group ; the largest is 1 corresponding to belonging to every cluster evenly (with probability $\frac{1}{T}$).

Overall, the entropy of memberships is low. The average entropy values per corpus are: 0.320 for PubMed (equivalent to belonging on average to 2-3 clusters), 0.324 for Twitter (2 clusters), 0.561 for Reddit (6-7 clusters) and 0.364 for Spotify (2-3 clusters). The low number of entities spread among clusters means that the clustering done by our model is easy to interpret --which eased the manual annotation of the clusters presented in the previous section.

\section{Conclusion}
In most previous approaches to information spreading, the effect of interactions between diffusing entities has been neglected. Here, we proposed a modified MMSBM that allows for investigating the role of interactions and quantifying them. By design, it also allows assessment of the importance of interactions compared with the virality of single pieces of information (i.e. their intrinsic ability to spread on their own). On this basis, we show that the effect of interactions on information spread is not trivial and that taking them into account increases predictive performance in several real-world applications. Following this observation, we proposed an implementation via a scalable EM algorithm allowing for its application on large datasets. 


However, a major limitation to our model is that it only accounts for pair interactions. While our conclusions state their importance in some real-world systems, it might not be sufficient for a number of others. For instance, a disease is seldom diagnosed on the basis of a single pair of symptoms; a model accounting for the interaction between n entities might be more relevant in this case. The same line of reasoning can be applied to gene and protein interactions \cite{SymptDiseNet}, bacteria symbiosis or even species co-evolution networks. We are currently working on a generalization of our model that will hopefully provide a better description of interacting processes at work everywhere in nature.

\bibliographystyle{ACM-Reference-Format}
\bibliography{references}
\appendix

\end{document}


\title[Supplementary Information: Information Interactions in Outcome Prediction]{Supplementary Information for: Information Interactions in Outcome Prediction: Quantification and Interpretation using Stochastic Block Models}

\author{Ga\"el Poux-M\'edard}
\affiliation{%
  \institution{Université de Lyon, Lyon 2, UR 3083}
  \streetaddress{5 avenue Pierre Mendès France}
  \city{Bron}
  \country{France}}
\email{gael.poux-medard@univ-lyon2.fr}

\author{Julien Velcin}
\affiliation{%
  \institution{Université de Lyon, Lyon 2, UR 3083}
  \streetaddress{5 avenue Pierre Mendès France}
  \city{Bron}
  \country{France}}
\email{julien.velcin@univ-lyon2.fr}

\author{Sabine Loudcher}
\affiliation{%
  \institution{Université de Lyon, Lyon 2, UR 3083}
  \streetaddress{5 avenue Pierre Mendès France}
  \city{Bron}
  \country{France}}
\email{sabine.loudcher@univ-lyon2.fr}

\renewcommand{\shortauthors}{Poux-M\'edard et al.}

\begin{abstract}
In most real-world applications, it is seldom the case that a result appears independently from an environment. In social networks, users’ behavior results from the people they interact with, news in their feed, or trending topics. In natural language, the meaning of phrases emerges from the combination of words. In general medicine, a diagnosis is established on the basis of the interaction of symptoms. Here, we propose the Interacting Mixed Membership Stochastic Block Model (IMMSBM), which investigates the role of interactions between entities (hashtags, words, memes, etc.) and quantifies their importance within the aforementioned corpora. We find that in inference tasks, taking them into account leads to average relative changes with respect to non-interacting models of up to 150\% in the probability of an outcome and greatly improves the predictions performances. Furthermore, their role greatly improves the predictive power of the model. Our findings suggest that neglecting interactions when modeling real-world phenomena might lead to incorrect conclusions being drawn.
\end{abstract}

\ccsdesc[500]{Information systems~Social recommendation}
\ccsdesc[500]{Information systems~Clustering}
\ccsdesc[300]{Mathematics of computing~Probabilistic inference problems}

\keywords{Interaction modeling, Block models, Information diffusion, Recommender systems}

\maketitle

\donotdisplay{
\section{Inference of the parameters}
We are now looking at inferring the model parameters, namely the components of the interactions tensor $ P_{i,j}(X_{i,j}=x)$. As stated in the previous section, we decomposed this tensor into an algebraic combination of the matrix $\theta$ and the tensor $p$. In this section, we introduce a method to infer them via an Expectation-Maximization (EM) algorithm. EM is a 2-step iterative algorithm. The first step consists of computing the expectation of the likelihood with respect to the latent variables, denoted $\omega_{i,j,x}(k,l)$, with parameters $\theta$, p set as constant. The second step consists of maximizing this expectation of the likelihood with respect to the parameters $\theta$, p. Iterating this process guarantees that the likelihood converges towards a local maximum.

\subsection{Expectation step}
Taking the logarithm of the likelihood as defined in Eq.\ref{eqLikelihood}, denoted $\ell$, we have:
\begin{equation}
\label{eqJensen}
    \begin{split}
         \ell &= \sum_{(i,j,x) \in R^{\circ}} \ln \sum_{k,l} \theta_{i,k}\theta_{j,l}p_{k,l}(x) \\
         &= \sum_{(i,j,x) \in R^{\circ}} \ln \sum_{k,l} \omega_{i,j,x}(k,l) \frac{\theta_{i,k}\theta_{j,l}p_{k,l}(x)}{\omega_{i,j,x}(k,l)} \\
         & \geq \sum_{(i,j,x) \in R^{\circ}} \sum_{k,l} \omega_{i,j,x}(k,l) \ln \frac{\theta_{i,k}\theta_{j,l}p_{k,l}(x)}{\omega_{i,j,x}(k,l)}
    \end{split}
\end{equation}
We used Jensen's inequality to go from the 2nd to 3rd line.
The inequality in Eq.\ref{eqJensen} becomes an equality for: \begin{equation}
\label{eqOmega}
    \omega_{i,j,x}(k,l) = \frac{\theta_{i,k}\theta_{j,l}p_{k,l}(x)}{\sum_{k',l'} \theta_{i,k'}\theta_{j,k'}p_{k', l'}(x)}
\end{equation}
where $\omega_{i,j,x}(k,l)$ is interpreted as the probability that the observation $(i,j,x)$ is due to i belonging to the group $k$ and j to $l$, that is the expectation of the likelihood of the observation $(i,j,x)$ with respect to the latent variables $k$ and $l$. Therefore, Eq.\ref{eqOmega} is the formula for the expectation step of the EM algorithm. An alternative derivation of the expectation step formula is discussed in \cite{Bishop}.

\subsection{Maximization step}
\label{maximization}
This step consists in maximizing the likelihood using the parameters of the model $\theta$ and p, independently of the latent variables.
In order to take into account the normalization constraints, we introduce the Lagrange multipliers $\phi$ and $\psi$. Following this, the constrained log-likelihood reads:
\begin{equation}
\label{eqLagMult}
    \ell_c = \ell - \sum_i (\phi_i \sum_{t} \theta_{i,t} - 1) - \sum_{k,l} (\psi_{k,l} \sum_{x} p_{k,l}(x) - 1)
\end{equation}
We first maximize $\ell_c$ with respect to each entry $\theta_{mn}$.
\begin{equation}
\label{eqDerivTheta}
    \begin{split}
        \frac{\partial \ell_c}{\partial \theta_{mn}} =& \frac{\partial \ell}{\partial \theta_{mn}} - \phi_m = 0\\
        =& \sum_{\partial m} \left( \sum_{l} \frac{\omega_{m,j,x}(n,l)}{\theta_{mn}} + \sum_{k} \frac{\omega_{j,m,x}(k,n)}{\theta_{mn}} \right) - \phi_m\\
        =& \frac{1}{\theta_{mn}}\sum_{\partial m} \left( \sum_{t} \omega_{m,j,x}(n,t) + \omega_{j,m,x}(t,n) \right) - \phi_m\\
        \Leftrightarrow\ \theta_{mn} =& \frac{1}{\phi_m}\sum_{\partial m} \sum_{t} \omega_{m,j,x}(n,t) + \omega_{j,m,x}(t,n)
    \end{split}
\end{equation}
where $\partial m$ stands for the set of observations in which the entry m appears $\{ (j,x) \vert (m,j,x) \in R^{\circ} \}$. Note that summing over this set in line 2 of Eq.\ref{eqDerivTheta} implies that the relation between two inputs entities is symmetric --if $(i,j,x) \in R^{\circ} \ \text{ implies } (j,i,x) \in R^{\circ}$. Summing the last line of Eq.\ref{eqDerivTheta} over $n \in T$ then multiplying it by $\phi_m$, we get an expression for $\phi_m$:
\begin{equation}
        \phi_m\sum_n \theta_{mn} = \phi_m = \sum_{\partial m} \sum_{t,n} \omega_{m,j,x}(n,t) + \omega_{i,m,x}(t,n) = \sum_{\partial m} 2 = 2 \cdot n_m
\end{equation}
Where $n_m$ is the total number of times that m appears as an input in $R^{\circ}$. Finally, plugging back this result in Eq.\ref{eqDerivTheta}, we get:
\begin{equation}
    \label{eqMaxTheta}
    \theta_{mn} = \frac{\sum_{\partial m} \left( \sum_{t} \omega_{m,j,x}(n,t) + \omega_{i,m,x}(t,n) \right)}{2.n_m}
\end{equation}

Following the same line of reasoning for p, we get:
\begin{equation}
    \label{eqMaxp}
    p_{r,s}(l) = \frac{\sum_{\partial l} \omega_{i,j,l}(r,s)}{\sum_{(i,j,l') \in R^{\circ}} \omega_{i,j,l'}(r,s)}
\end{equation}
The set of Equations \ref{eqMaxTheta} and \ref{eqMaxp} constitutes the maximization step of the EM algorithm. They hold only if the input entities interactions are symmetric (e.g. when $\{ (j,x) \vert (m,j,x) \in R^{\circ} \} = \{ (i,x) \vert (i,m,x) \in R^{\circ} \}$), which is what we aimed to do.
It is worth noting that the proposed algorithm offers linear complexity with the size of the dataset $\mathcal{O}(\vert R^{\circ} \vert)$ provided the number of clusters is constant, while guaranteeing convergence to a local maximum.

\donotdisplay{
}

\section{Choice of model and protocol}
The only parameter to be set in our model is the number of clusters. To determine it, we run experiments on each of the datasets with various numbers of clusters T, ranging from 5 to 50 by step of 5. Then, we select the number of clusters minimizing the Akaike's information criterion (AIC) \cite{Akaike1973,refChoixAIC}, which coincides with a lesser improvement of the model's performances with increasing T. Figures illustrating the role of the number of clusters in model's performance are provided in SI, Section~3; they support the AIC-based choice of the number of clusters, as it coincides with the number after which performance does not improve significantly with T anymore.

Once the model is selected, we perform 100 independent runs, each with independent random initialization of the parameters $\theta$ and p. The EM loop stops once the relative variation of the likelihood between two iterations is less than 0.001\%.

\donotdisplay{
\begin{figure}
    \includegraphics[width=.25\textwidth]{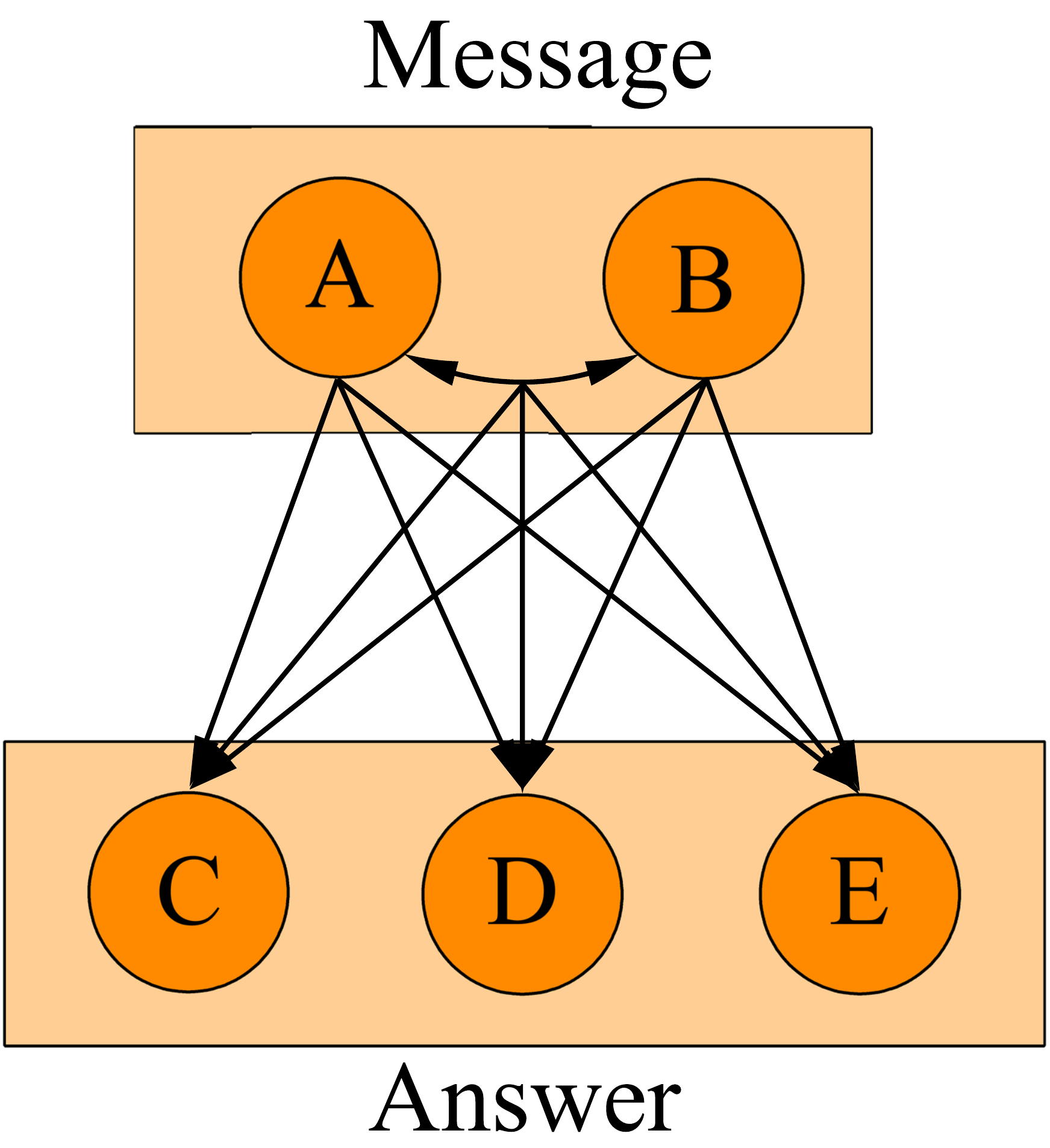} 
    \caption{\textbf{Datasets building} - Illustration of the construction of the triplet datasets. We take every pair of entities AA, BB, CC, AB, AC, BC in a message (named entities in a Reddit post, symptoms in a medical report, etc.) and couple them to every entity D, E, F in the answer (named entities in a Reddit answer, diseases in a medical report, etc.). Self-pairs (AA, BB, CC) allow access to the intrinsic virality of A, B and C.} \label{figConstrCorp}
\end{figure}
}

\section{Well-calibrated predictions}
\label{ProbCal}
\begin{figure}[h]
    \includegraphics[width=0.7\textwidth]{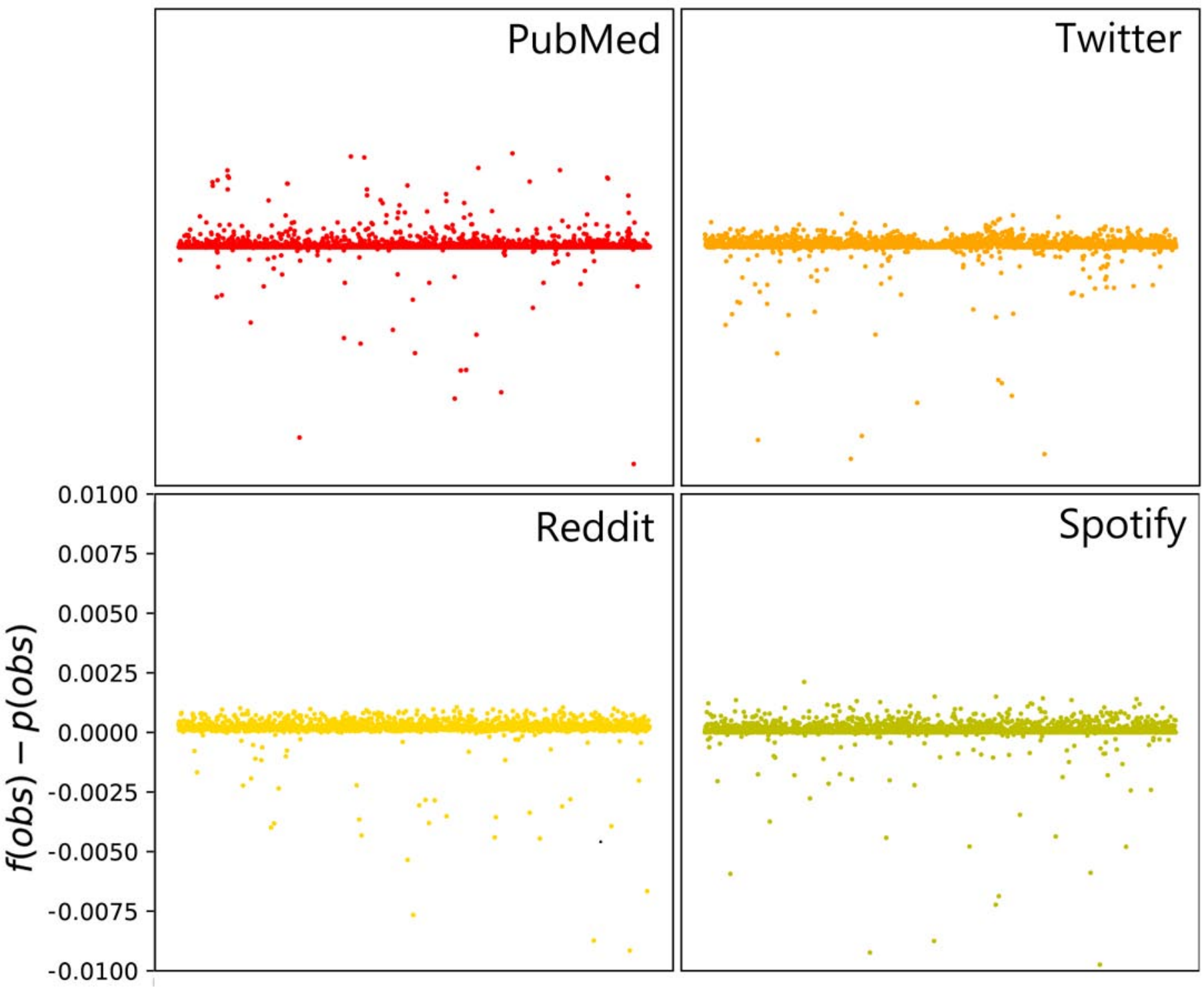}
    \caption{\textbf{Probabilistic calibration} - Difference between the frequency of an observed entity $f(obs)$ minus its probability given the model, $p(obs)$ on the y-axis. The corresponding observations are distributed along the x-axis. The difference being very small, it means our model achieves to capture the stochastic processes at stake in the datasets. The average of the absolute difference for each of the datasets is: $2.10^{-4}$ (PubMed), $3.10^{-4}$ (Twitter), $5.10^{-4}$ (Reddit) and $3.10^{-4}$ (Spotify)} \label{figCalib}
\end{figure}
We first evaluate the probabilistic calibration of our predictions. Although the original article introducing this measure consider the case of continuous probability distributions ouputs, the transposition to the discrete case (multilabel classification) is possible \cite{Calibration,AntoniaScalableRS}. The measure consists in comparing the probabilistic output distribution of an observation yielded by a model ($p(obs)$) with the actual frequency of the outputs witnessed in the test-set ($f(obs)$). The smaller the difference, the better the model captures the hidden patterns of the data generative process. When the output data is generated by a process that does not rely on the combination of pairs of outputs, it is likely that the probabilistic predictions made by IMMSBM does not account for the actual frequency of the outputs. 

We present these results in Fig.\ref{figCalib} for every dataset. As we can see, the difference between observed frequency and modeled probability of an output is very small. The negligible gap between those quantities for the IMMSBM therefore means the model correctly accounts for the stochastic mechanisms between the publication of an ouput given inputs. In other words, this measure assesses the relevancy of modelling these data via an interacting stochatsic block model. This adds to the robustness of our model, which correctly accounts for real-world processes. Therefore, one can interpret the model's parameters as a good approximation of stochastic processes at stake in the considered datasets. 

}

\section{Datasets}
\label{datasets}
\subsection{Medical records}
The Pubmed dataset collect has been inspired by \cite{SymptDiseNet}. Every article on PubMed is manually annotated by experts with a list of keywords describing the main topics of the publication. We downloaded a list of 322 symptoms and 4,442 diseases provided by \cite{SymptDiseNet}. Then, we used the PubMed API to query each one of the symptom/disease keywords aforementioned. For each result, we got a list of every publication in which the keyword is among the main topics. Then, we build the dataset by considering every publication in which there is at least one symptom and one disease. Finally, we create the triplets (symptom1, symptom2, disease) by looking at all the pairs of symptoms in an article and linking each one of them to all the diseases observed in the same article. In the end, we are left with a total of 52,833,690 observed triplets, distributed over 15,809,271 PubMed publications.

\subsection{Spotify}
The Spotify dataset has been collected using the Spotify API. We randomly sampled 2,000 playlists using the keywords "english" and "rock", which corresponds to a total of 135,100 songs. Then, for each playlist, we used a running window of 4 songs to build the dataset. The artist of the song immediately after the running window is the output we aim at predicting, x, and the artists of the 4 songs within the running window are the interacting inputs. Once again, we consider all the possible pairs of artists in the running window and associate them to the output artist x. Note that we only considered the artists that appear more than 50 times in the whole dataset, for the sake of statistical relevance. The resulting dataset is consists of 1,236,965 triplets for 2,028 artists.

\subsection{Twitter}
We gathered the Twitter dataset used in \cite{DataSetTwitter}. It consists in a collection of all the tweets containing URLs posted during the month of October 2010. A first operation consisted in cleaning the dataset of URLs that are considered as aggressive advertising. To do so, we considered only the URLs whose retweets built a chain of length at least 50; this choice comes from the idea that commercial spams are not likely to be retweeted by actual users and therefore do not create chains. Secondly, we considered only the users who have not tweeted a given URL more than 5 times, this behavior being an activity typical of spamming bots. Doing so, we are left with tweets that are mostly coming from the non-commercial activity of human users. Then, we follow a dataset building process similar to \cite{CoC}. For each user, we slice her feed + tweets temporal sequence in intervals separated by the tweets of the user. Every time a user tweets something, the interval ends. An interval therefore consists of the tweet of the user and all the tweets she has been exposed to right before tweeting. Following the suggestion of \cite{CoC}, we only consider the 3 last tweets the user has been exposed to before retweeting one of them. Each one of these intervals form an entry of our message+answer dataset (3 last entries in the feed + next tweets). In the end, we are left with 284,837 intervals containing a total of 2,110 different tweeted URLs. Our dataset then consists of 1,181,543 triplets.

\subsection{Reddit}
Finally, we downloaded the May 2019 Reddit dataset from the data repository pushhift.io, which stores regular saves of the comments posted on the website. We chose to consider only the comments made in the subreddit r/news. Reddit's comments system work as a directed tree network, where each answer to a given comment initiates a new branch. We considered pairs of messages such as one (the answer) has for direct parent the other one (the comment). We then extracted all the named entities in both of them using the Spacy Python library. For each pair of named entities in the comment, we associated every name entity in the answer. We consider here only the named entities that appear at least 200 times in the subreddit, for the same reason as for the Spotify dataset. The final dataset results in 35,364,725 triplets for a total of 1,656 named entities.

\donotdisplay{

\section{Results as a function of the number of clusters}
\begin{figure}[h]
    \includegraphics[width=1.0\textwidth]{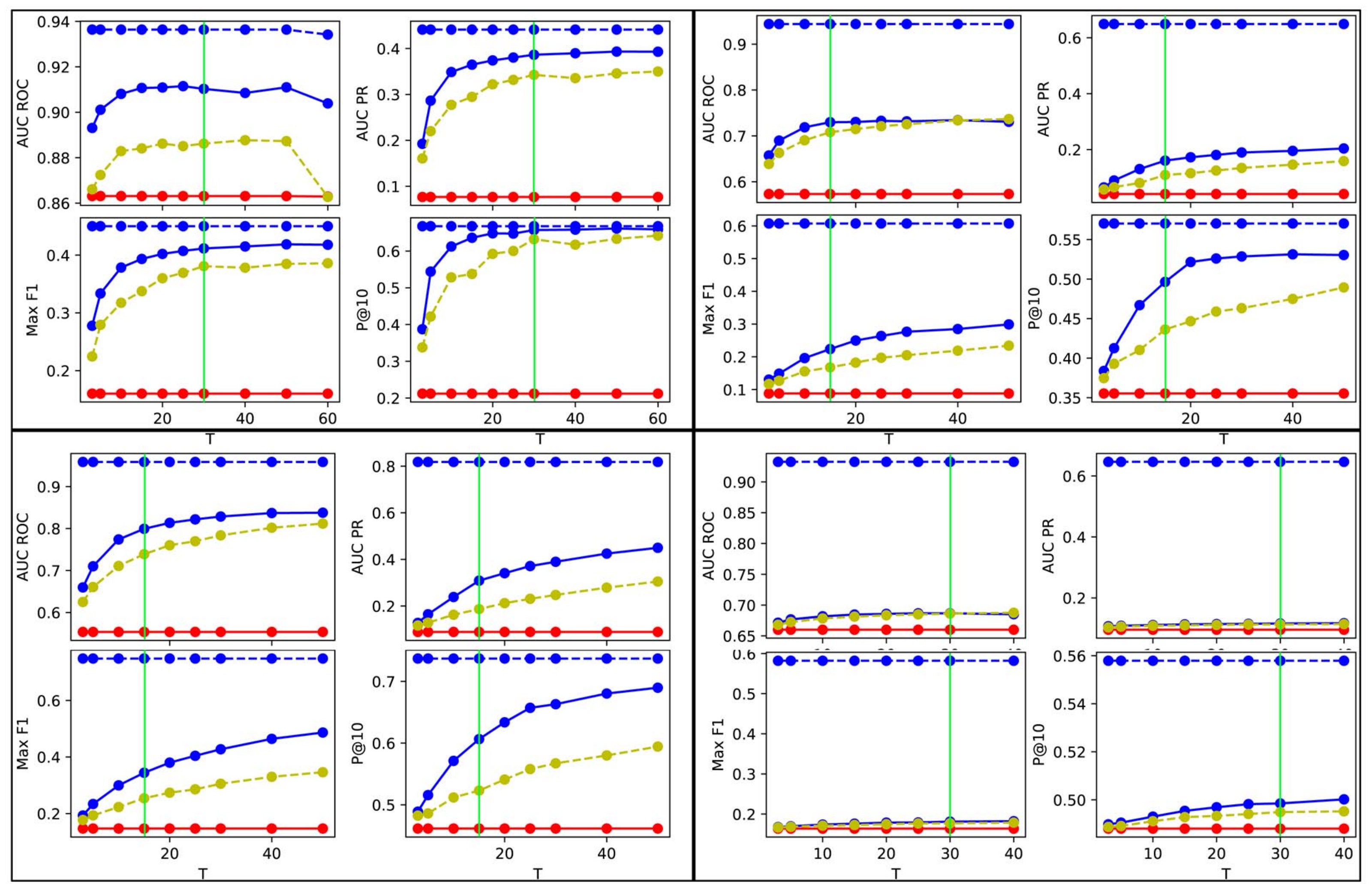}
    \caption{Performance variations on all the metrics for every dataset considered. Dashed blue line: upper limit to performances ; blue line: IMMSBM ; yellow dashed line: MMSBM ; red line: naive baseline. Top left: PubMed ; top right: Spotify ; bottom left: Twitter ; bottom right: Reddit. The vertical green line shows the selected number of clusters.} \label{figCalib}
\end{figure}
}

\section{Upper limit to predictions}
\label{upperLim}
We derive an analytical expression for the upper-limit to our model for a given dataset. Explicitly, we analytically maximize the likelihood according to each entry of the dataset.

We enforce the constraint that the sum over the output space of probabilities given any observations made has to sum to 1. To do so, we use a Lagrange multiplier $\lambda_{obs}$ for every different observation (in the case of our model: for every different triplet). The log-likelihood then takes the following form:
\begin{equation}
    \label{eqLimHauteP}
    \begin{split}
        &\ell = \sum_{(obs, x)} \ln P_{obs}(x) - \sum_{obs}\lambda_{obs}(\sum_x P_{(obs,x)}-1)\\
        \Leftrightarrow\ &\frac{\partial \ell}{\partial P_{obs_i}(x_i)} = \sum_{\partial (obs_i,x_j)}\frac{1}{P_{obs_i}(x_j)} - \lambda_{obs_i} = 0 \\
        \Leftrightarrow\ &P_{obs_i}(x_j) = \frac{1}{\lambda_{obs_i}}\sum_{\partial (obs_i,x_j)} 1
    \end{split}
\end{equation}
Where $obs_i$ correspond to any given couple of inputs (i,j) in the model presented in the main paper. We use the following notation: $\partial (obs_i,x_j) = \{ obs \vert (obs_i, x_j) \in R^{\circ} \}$, with $R^{\circ}$ the dataset entries. Therefore, we can define $\sum_{\partial (obs_i,x_j)} 1 \equiv N_{obs_i,x_j}$ the number of times $obs_i$ appears jointly with $x_j$ in the dataset. We are now looking for the $\lambda_{obs_i}$ maximizing the likelihood:
\begin{equation}
    \label{eqLimHauteLamb}
    \begin{split}
        &\frac{\partial \ell}{\partial \lambda_{obs_i}} = \sum_x P_{obs_i}(x)-1  = 0 \\
        \Leftrightarrow\ &\sum_x \frac{N_{obs_i,x}}{\lambda_{obs_i}} = 1 \\
        \Leftrightarrow\ &\lambda_{obs_i} = \sum_x N_{obs_i,x}\\
    \end{split}
\end{equation}
Finally, plugging Eqs.\ref{eqLimHauteP} and \ref{eqLimHauteLamb} together, we obtain:
\begin{equation}
    P_{obs_i}(x_j) = \frac{N_{obs_i, x_j}}{\sum_x N_{obs_i, x}}
\end{equation}
This equation gives the probability maximizing the likelihood for any entry of the dataset. In the main model, it translates to $P_{(i,j)}(x) = N_{(i,j), x}/\sum_x' N_{(i,j), x'}$ with $N_{(i,j),x}$ the number of times output x has been witnessed after a pair of inputs (i,j). Note that this is simply the frequency of an output given a pair of input entities.

Keep in mind that the result we just derived gives perfect predictions only for a particular dataset, and therefore and has no global predictive value. This tool is useful when it comes to assess the performance of other models' results, but is unusable in prediction tasks.

\donotdisplay{

\section{Clusters composition}
We provide in Tables \ref{TabPubMed},\ref{TabRed} and \ref{TabSpot} the exhaustive list of entities belonging to each cluster with more than 50\% membership (80\% for Spotify) for the datasets considered. 

Each one of the clusters has been manually given a name according to the lists present in the tables. The names presented here are the same as the ones shown in Fig.3 of the main article.
\begin{table*}
\caption{Composition of clusters for the PubMed dataset. \label{TabPubMed}}
\centering
\setlength{\lgCase}{12cm}
\begin{tabular}{|p{0.2\lgCase}|p{\lgCase}|p{0\lgCase}}
 \cline{1-2}
 \centering Cluster & \centering Entities &  \\
 \cline{1-2}
1 - Vision & anisocoria, blindness, color vision defects, diplopia, eye hemorrhage, eye manifestations, eye pain, miosis, photophobia, pseudophakia, pupil disorders, scotoma, sensation disorders, tonic pupil, usher syndromes, vision, vision disorders, waterhouse-friderichsen syndrome & \\ 
 \cline{1-2} 
 2 - Blood & ecchymosis, glossalgia, oral hemorrhage, purpura & \\ 
 \cline{1-2} 
 3 - Seizures & seizures & \\ 
 \cline{1-2} 
 4 - Neurons & neurologic manifestations & \\ 
 \cline{1-2} 
 5 - Male genitals & dysmenorrhea, dysuria, encopresis, hirsutism, pelvic pain, prostatism, urinary bladder, virilism & \\ 
 \cline{1-2} 
 6 - Eating dis. & hyperphagia, obesity, overweight & \\ 
 \cline{1-2} 
 7 - Coma & coma, consciousness disorders, hypothermia, tetany, unconsciousness & \\ 
 \cline{1-2} 
 8 - Pain & pain & \\ 
 \cline{1-2} 
 9 - Facial pain & facial pain, toothache, trismus & \\ 
 \cline{1-2} 
 10 - Back pain & back pain, hot flashes, low back pain & \\ 
 \cline{1-2} 
 11 - Muscle & muscle cramp, muscle hypertonia, muscle rigidity, muscle spasticity, muscle weakness, myotonia & \\ 
 \cline{1-2} 
 12 - Mental & cafe-au-lait spots, deaf-blind disorders, hydrops fetalis, mental retardation, morning sickness & \\ 
 \cline{1-2} 
 13 - Paralysis & paralysis & \\ 
 \cline{1-2} 
 14 - Undefined & neuralgia, piriformis muscle syndrome & \\ 
 \cline{1-2} 
 15 - Speech disorder & articulation disorders, communication disorders, language disorders, speech disorders & \\ 
 \cline{1-2} 
 16 - Kidney & albuminuria, flank pain, hemoglobinuria, oliguria, polyuria, proteinuria, renal colic & \\ 
 \cline{1-2} 
 17 - Voice & aphonia, dysphonia, hoarseness, orthostatic intolerance, respiratory aspiration, vocal cord paralysis, voice disorders & \\ 
 \cline{1-2} 
 18 - Neuropsychatry & agraphia, alien hand syndrome, aphasia, gerstmann syndrome, headache, hemiplegia, paresis, systolic murmurs & \\ 
 \cline{1-2} 
 19 - Sceptisemia & hypergammaglobulinemia, pruritus, purpura fulminans, tinea pedis & \\ 
 \cline{1-2} 
 20 - Vertigo & dizziness, earache, hearing disorders, hyperacusis, motion sickness, presbycusis, tinnitus, vertigo & \\ 
 \cline{1-2} 
 21 - Mental & agnosia, alexia, anomia, auditory perceptual disorders, bulimia, delirium, dyslexia, hallucinations, learning disorders, memory disorders, mutism, neurobehavioral manifestations, perceptual disorders, phantom limb, pseudobulbar palsy, stuttering & \\ 
 \cline{1-2} 
 22 - Articulations & arthralgia, metatarsalgia, shoulder pain & \\ 
 \cline{1-2} 
 23 - Muscle weight & birth weight, body weight, labor pain, sarcopenia & \\ 
 \cline{1-2} 
 24 - Heart & acute coronary syndrome, amaurosis fugax, angina, angina pectoris, chest pain, heart murmurs, syncope & \\ 
 \cline{1-2} 
 25 - Liver & hyperemesis gravidarum, jaundice, necrolytic migratory erythema & \\ 
 \cline{1-2} 
 26 - Breath & cough, dyspnea, hemoptysis, respiratory paralysis & \\ 
 \cline{1-2} 
 27 - Sleep & sleep deprivation, sleep disorders, snoring & \\ 
 \cline{1-2} 
 28 - Abdominal pain & abdomen, abdominal pain, diarrhea, dyspepsia, eructation, flatulence, gastroparesis, vomiting & \\ 
 \cline{1-2} 
 29 - Movement & amblyopia, ataxia, athetosis, catalepsy, cerebellar ataxia, chorea, dyskinesias, dystonia, hyperkinesis, hypokinesia, myoclonus, supranuclear palsy, tics, torticollis, tremor & \\ 
 \cline{1-2} 
 30 - Brain & ageusia, amnesia, cerebrospinal fluid otorrhea, cerebrospinal fluid rhinorrhea, decerebrate state, persistent vegetative state & \\ 
 \cline{1-2} 
 
 \cline{1-2}
\end{tabular}
\end{table*}

\begin{table*}
\caption{Composition of clusters for the Reddit dataset. \label{TabRed}}
\centering
\setlength{\lgCase}{12cm}
\begin{tabular}{|p{0.2\lgCase}|p{\lgCase}|p{0\lgCase}}
 \cline{1-2}
 \centering Cluster & \centering Entities &  \\
 \cline{1-2}
1 - Komodo dragon & dragon, eye, king, komodo, link, mate, ring & \\ 
 \cline{1-2} 
 2 - Diseases & antivaxxer, autism, child, disease, gene, herpes, immunity, measles, outbreak, parent, vaccination, vaccine & \\ 
 \cline{1-2} 
 3 - Weapons & ar15, felon, firearm, gun, handgun, knife, nra, nz, ownership, pistol, rifle, shooter, shooting, shotgun, weapon, zealand & \\ 
 \cline{1-2} 
 4 - Driving & accident, car, driver, driving, drunk, dui, license, road, spider, traffic, vehicle & \\ 
 \cline{1-2} 
 5 - Media & click, cnn, coverage, forum, fox, karma, marathon, media, meme, motive, news, outlet, page, post, reddit, sub, subreddit, td, upvote & \\ 
 \cline{1-2} 
 6 - J. Smollet hoax & hoax, jussie, maga, smollett, supporter & \\ 
 \cline{1-2} 
 7 - Punishment & capital, cruel, death, execution, inmate, murderer, offender, penalty, prison, prisoner, punishment, rehabilitation, revenge, row, sentence, torture & \\ 
 \cline{1-2} 
 8 - Catholicism & aid, bible, catholic, church, conversion, gay, homosexuality, marriage, mormon, pedophile, priest, satan, scout, temple, therapy, troop & \\ 
 \cline{1-2} 
 9 - Planes & air, airline, airport, battery, boeing, bottle, delivery, flight, passenger, pilot, plane, seat, tip, tsa & \\ 
 \cline{1-2} 
 10 - Statistics & average, cdc, homicide, increase, math, number, percent, rate, stat, statistic, study, suicide & \\ 
 \cline{1-2} 
 11 - Managment & business, ceo, charity, corporation, economy, employee, executive, fund, incentive, income, investment, market, pay, payer, profit, raise, revenue, salary, shareholder, spending, stock, tax, wage, wealth & \\ 
 \cline{1-2} 
 12 - Undefined & fallacy, slavery & \\ 
 \cline{1-2} 
 13 - Animals & animal, bag, beef, burger, cat, chicken, cow, dog, dumpster, factory, farm, farmer, food, fry, meat, milk, pet, pig, pit, plastic, poacher, puppy, restaurant, taste, vegan & \\ 
 \cline{1-2} 
 14 - Notre-Dame fire & art, building, cathedral, construction, dame, design, fire, france, glas, lead, notre, pari, roof, spire, stone, structure, tower & \\ 
 \cline{1-2} 
 15 - US politics & bernie, bush, campaign, candidate, clinton, congres, dem, democrat, election, gop, governor, hillary, obama, party, patriot, president, republican, senator, vote, voter & \\ 
 \cline{1-2} 
 16 - Police & cop, department, officer, polouse, taser, union & \\ 
 \cline{1-2} 
 17 - Feminism & birth, consent, discrimination, equality, feminism, feminist, gap, gender, incel, male, partner, peni, pill, sexism, tran, woman & \\ 
 \cline{1-2} 
 18 - School & bathroom, bully, clas, class, college, district, grade, school, student, teacher, university & \\ 
 \cline{1-2} 
 19 - US laws & amendment, constitution, freedom, liberty, ruling, speech, supreme, violation & \\ 
 \cline{1-2} 
 20 - International & arabia, china, east, eu, iran, iraq, oil, regime, saudi, socialism, tank, venezuela, war & \\ 
 \cline{1-2} 
 21 - Medicine & addict, addiction, cure, doctor, drug, fentanyl, healthcare, hospital, insulin, insurance, med, medication, medicine, patient, pharma, professional, substance, surgery, tooth, treatment, va & \\ 
 \cline{1-2} 
 22 - GAFA & amazon, apple, artist, computer, device, facebook, password, phone, software, tech, technology, thief & \\ 
 \cline{1-2} 
 23 - Smokers & capacity, cigarette, corner, disney, pack, park, smell, smoke, smoker, smoking, weed & \\ 
 \cline{1-2} 
 24 - Immigration & asylum, border, citizenship, entry, felony, illegal, immigrant, immigration, mexico, migrant, militia, patrol, port & \\ 
 \cline{1-2} 
 25 - Trial & accusation, attorney, bail, charge, client, condom, da, judge, jury, lawyer, plea, precedent, prosecution, prosecutor, rape, sweden, trial & \\ 
 \cline{1-2} 
 26 - Extremisms & alt, antifa, antisemitism, atheist, christian, christianity, extremist, fascism, fascist, hitler, ideology, islam, islamic, israel, jew, kkk, mosque, muslim, nationalism, nationalist, nazi, religion, supremacist, supremacy, symbol, terrorism, terrorist & \\ 
 \cline{1-2} 
 27 - Wikileak & alright, assange, attitude, dirt, dnc, email, info, intelligence, journalist, leak, proof, propaganda, public, putin, russia, russian, support, west, wikileak & \\ 
 \cline{1-2} 
 28 - US Coasts & apartment, area, baltimore, chicago, city, coast, francisco, homeles, housing, la, neighborhood, nyc, poop, san, sf, shelter, shithole, south, town, weather, york & \\ 
 \cline{1-2} 
 29 - Hobbies & band, channel, character, club, cult, episode, festival, game, golf, hitman, movie, music, night, pizza, player, podcast, robert, season, series, tiger, watch & \\ 
 \cline{1-2} 
 30 - Students & age, beer, card, credit, debt, drink, drinking, loan, minimum & \\ 
 \cline{1-2}

 \cline{1-2}
\end{tabular}
\end{table*}

\begin{table*}
\caption{Composition of clusters for the Spotify dataset. Note that here we define the membership threshold at 80\%: since T is smaller, more entities are likely to belong to any cluster more than 50\%, hence leading to hardly human-readable results. \label{TabSpot}}
\centering
\setlength{\lgCase}{12cm}
\begin{tabular}{|p{0.25\lgCase}|p{\lgCase}|p{0\lgCase}}
 \cline{1-2}
 \centering Cluster & \centering Entities &  \\
 \cline{1-2}
1 - Latino & alaska y dinarama, alejandro sanz, andrés calamaro, arena hash, asian kung-fu generation, aterciopelados, bacilos, café tacvba, caifanes, d.a.n., donots, duncan dhu, ekhymosis, el gran silencio, el tri, elefante, fito paez, fobia, frankie valli and the four seasons, gustavo cerati, hawkwind, heroes del silencio, hombres g, jaguares, jarabe de palo, jethro tull, jr jr, jumbo, la ley, la mosca tse-tse, la unión, los bunkers, los claxons, los enanitos verdes, los fabulosos cadillacs, los prisioneros, los tres, magneto, maldita vecindad y los hijos del 5to. patio, maná, mar de copas, marty robbins, mikel erentxun, molotov, mägo de oz, nacha pop, nek, paellas, pedro suárez-vértiz, rata blanca, soda stereo, the horrors, van der graaf generator, victimas del doctor cerebro, vilma palma e vampiros & \\ 
 \cline{1-2} 
 2 - Blues rock & cartel de santa, david gilmour, hank williams, johnny hallyday, keith richards, kiss, melbourne symphony orchestra, myles kennedy & \\ 
 \cline{1-2} 
 3 - Christian rock \& Alt. rock & 6cyclemind, angee rozul, audio adrenaline, bamboo, beyond creation, charles bradley, cheese, chicosci, chris quilala, cueshé, eraserheads, francism, franco, hale, hilera, hillsong united, imago, kamikazee, kjwan, mayra andrade, mosaic msc, newsboys, orange and lemons, parokya ni edgar, petra, powerspoonz, rico blanco, rivermaya, sandwich, shawn mcdonald, slapshock, static-x, stray kids, the afters, the dawn, the speaks, the stone foxes, the twang, tobymac, typecast, urbandub & \\ 
 \cline{1-2} 
 4 - Metal & a skylit drive, fort minor, heilung, limp bizkit, p.o.d., papa roach, powerwolf & \\ 
 \cline{1-2} 
 5 - J-pop \& Alt. rock & alison mosshart, bigbang, bring me the horizon, crossfaith, dani filth, egoist, exo, exo-k, grimes, hyde, lisa, one ok rock, placebo, primal scream, reol, self deception, the oral cigarettes, vamps, ves tal vez, vixx, while she sleeps & \\ 
 \cline{1-2} 
 6 - Electro \& House & afrojack, diplo, dragonland, gloria trevi, james bay, labrinth, lost frequencies, lsd, nayer, scouting for girls, the word alive, walk off the earth & \\ 
 \cline{1-2} 
 7 - New wave & alphaville, baltimora, bananarama, barbra streisand, belinda carlisle, billy ocean, bronski beat, cool kids of death, culture club, cyndi lauper, desireless, dreamtale, erasure, eric 'et' thorngren, fine young cannibals, gentle giant, googolplex, haschak sisters, huey lewis and the news, imagination, information society, johnny hates jazz, kool and the gang, laura branigan, level 42, men at work, michael sembello, opeth, orchestral manoeuvres in the dark, pet shop boys, roxy music, soft cell, spandau ballet, starship, tears for fears, the bangles, the buggles, the human league, the j. geils band, the romantics, thompson twins, wang chung & \\ 
 \cline{1-2} 
 8 - Rock'n'roll \& J.Lennon & bill haley and his comets, chuck berry, frankie valli, jay and the americans, john denver, john lennon, little richard, the beach boys, the flux fiddlers, the plastic ono band, van morrison & \\ 
 \cline{1-2} 
 9 - Undefined & jeff buckley, karnivool, nirvana & \\ 
 \cline{1-2} 
 10 - Heavy metal & anthrax, arsenal, avenged sevenfold, beatsteaks & \\ 
 \cline{1-2} 
 11 - Folk rock & coldplay, dark moor, dermot kennedy, james blunt, power quest, taburete, tom walker & \\ 
 \cline{1-2} 
 12 - Recent pop & alan walker, alessia cara, alesso, alexandra porat, aloe blacc, anne-marie, arizona zervas, au/ra, ava max, axwell /\ ingrosso, becky hill, before you exit, benny blanco, billie eilish, blackbear, blackpink, bloodpop®, bonn, bryson tiller, camila cabello, cardi b, chance the rapper, clean bandit, daddy yankee, dakota, daya, demi lovato, digital farm animals, disciples, dj khaled, dj snake, elina, farruko, fifth harmony, francesco yates, grey, hailee steinfeld, halsey, illenium, jack and jack, jax jones, jessie reyez, jonas blue, jonas brothers, jp cooper, julia michaels, justin bieber, k-391, kane brown, khalid, kygo, lauv, lennon stella, liam payne, little mix, luis fonsi, major lazer, maren morris, marshmello, martin garrix, mø, niall horan, nick jonas, noah cyrus, normani, sabrina carpenter, sandro cavazza, selena gomez, shawn mendes, steeleye span, stefflon don, tainy, tiësto, tomine harket, troye sivan, why don't we, william singe, zara larsson, zedd & \\ 
 \cline{1-2} 
 13 - Alt. rock & blutengel, brian eno, greta van fleet, ledger, mew, muse, starsailor & \\ 
 \cline{1-2} 
 14 - Pop/Glam-rock & andy black, asobi seksu, el haragán y compañía, funeral for a friend, panic! at the disco, suspekt, twenty one pilots & \\ 
 \cline{1-2} 
 15 - Garage rock & interpol, kings of leon, liquido, mono, rainbow99, the hives & \\ 
 \cline{1-2} 
\end{tabular}
\end{table*}

\donotdisplay{

\section{Graphical schema of the model}
\begin{figure}[H]
    \centering
    
    \begin{tikzpicture}
    \tikzstyle{main}=[circle, minimum size = 8mm, thick, draw =black!80, node distance = 13mm]
    \tikzstyle{connect}=[-latex, thick]
    \tikzstyle{box}=[rectangle, draw=black!100]
      \node[main, fill = black!20] (alpha1) [label=center:$i$] { };
      \node[main, fill = black!20] (alpha2) [right=5mm of alpha1,label=center:$j$] { };
      \coordinate (MiddleAlpha) at ($(alpha1)!0.5!(alpha2)$);
      \node[main] (t1) [below=8mm of alpha1,label=center:$k$] { };
      \node[main] (t2) [below=8mm of alpha2,label=center:$l$] { };
      \coordinate (Middle) at ($(t1)!0.5!(t2)$);
      \node[main, fill = black!20] (z)  [below=12mm of Middle,label=center:$x$] { };
      
      \node[main] (theta) [right=1.cm of t2,label=center:$\Vec{\theta}$] { };
      \node[main] (pkl) [right=1.67cm of z,label=center:$\Vec{p}$] { };
      
    
      \path 
            (alpha1) edge [connect] (t1)
            (alpha2) edge [connect] (t2)
            (t1) edge [connect] (z)
            (t2) edge [connect] (z)
            
    		(theta) edge [bend left] [connect] (t1)
    		(theta) edge [connect] (t2)
    		(pkl) edge [connect] (z);
    		
      
      \node[rectangle, inner sep=0mm,label=below:$R^{\circ}$, xshift=11.5mm, yshift=1mm,  fit= (z) (alpha1) (alpha2) (t1) (t2)] {};
      \node[rectangle, inner sep=3.4mm,draw=black!100,  fit= (z) (alpha1) (alpha2) (t1) (t2)] {};
      

      \node[rectangle, inner sep=0mm, fit= (theta),label=below:$T$, yshift=0.7mm,xshift=4.7mm] {};
      \node[rectangle, inner sep=3.2mm,draw=black!100, fit= (theta)] {};
      
      \node[rectangle, inner sep=0mm, fit= (pkl),label=below:$T \times T$, yshift=0.7mm,xshift=2mm] {};
      \node[rectangle, inner sep=3.2mm,draw=black!100, fit= (pkl)] {};
      
    \end{tikzpicture}
    \caption{Graphical representation of the model. In the generation of an output, two entities are randomly selected from the input space of a message. Then for each one of them we draw a cluster from a Dirichlet distribution of parameter $\alpha$, encoded in the $\theta$ matrix. Finally, looking at the interaction between the clusters, we generate an output x from a multinomial distribution of parameter $\beta$, encoded in the cluster interaction matrix p.}
\end{figure}
}
}

\bibliographystyle{ACM-Reference-Format}
\bibliography{sample-base}
\appendix